\documentclass{ieeeaccess}
\usepackage{cite}
\usepackage{amsmath,amssymb,amsfonts}
\usepackage{graphicx}
\usepackage{textcomp}
\def\BibTeX{{\rm B\kern-.05em{\sc i\kern-.025em b}\kern-.08em
    T\kern-.1667em\lower.7ex\hbox{E}\kern-.125emX}}

\usepackage{float} % Optional, to use [H] float option
\usepackage{siunitx}
\usepackage[margin=1in]{geometry}
\usepackage{textcomp}
\usepackage{multirow}
\usepackage{listings}
\usepackage{xcolor}
\usepackage{booktabs}
\usepackage{natbib}
\usepackage{breakcites}
\usepackage{comment}
\usepackage{array}
\usepackage{tabularx} % For more flexible tables
\usepackage{booktabs} % For better-looking tables
\usepackage{caption} % For customizing captions
\usepackage{algpseudocode}
\usepackage{amsmath}
\usepackage{tabularx} % For adjustable column widths
\usepackage{caption} % For better caption control
\usepackage{booktabs}
\usepackage{algorithm}
\usepackage{amsmath}
\usepackage{placeins}
\DeclareMathOperator*{\argmax}{arg\,max}
\usepackage{cite}
\usepackage{lipsum}

\begin{document}
\history{Date of publication October 2024, date of current version October 18, 2024.}
\doi{10.1109/ACCESS.2024.3484663}

\title{CG-CNN: Self-Supervised Feature Extraction Through Contextual Guidance and Transfer Learning}

\author{\uppercase{Olcay Kursun}\authorrefmark{1}, 
\uppercase{Ahmad Patooghy\authorrefmark{2}, \IEEEmembership{Senior, IEEE},  Peyman Poursani}\authorrefmark{2}, \uppercase{Oleg V. Favorov}\authorrefmark{3}}

\address[1]{Department of Computer Science, Auburn University at Montgomery, AL 36117, USA  (e-mail: okursun@aum.edu)}
\address[2]{Department of Computer Systems Technology, North Carolina A\&T State University, Greensboro, NC 27411, USA (e-mail: apatooghy@ncat.edu, pjpoursani@aggies.ncat.edu)}
\address[3]{Joint Department of Biomedical Engineering, University of North Carolina at Chapel Hill,
NC, 27599 USA (e-mail: favorov@email.unc.edu)}
%\tfootnote{???This paragraph of the first footnote will contain support  information, including sponsor and financial support acknowledgment. For  example, ``This work was supported in part by the U.S. Department of  Commerce under Grant BS123456.''}

\markboth
{Kursun \headeretal: Accepted to IEEE ACCESS DOI: 10.1109/ACCESS.2024.3484663}
{Kursun \headeretal: Accepted to IEEE ACCESS DOI: 10.1109/ACCESS.2024.3484663}

\corresp{Corresponding author: Ahmad Patooghy (e-mail: apatooghy@ncat.edu).}

\begin{abstract}
Contextually Guided Convolutional Neural Networks (CG-CNNs) employ self-supervision and contextual information to develop transferable features across diverse domains, including visual, tactile, temporal, and textual data. This work showcases the adaptability of CG-CNNs through applications to various datasets such as Caltech and Brodatz textures, the VibTac-12 tactile dataset, hyperspectral images, and challenges like the XOR problem and text analysis. In text analysis, CG-CNN employs an innovative embedding strategy that utilizes the context of neighboring words for classification, while in visual and signal data, it enhances feature extraction by exploiting spatial information. CG-CNN mimics the context-guided unsupervised learning mechanisms of biological neural networks and it can be trained to learn its features on limited-size datasets. Our experimental results on natural images reveal that CG-CNN outperforms comparable first-layer features of well-known deep networks such as AlexNet, ResNet, and GoogLeNet in terms of transferability and classification accuracy. In text analysis, CG-CNN learns word embeddings that outperform traditional models like Word2Vec in tasks such as the 20 Newsgroups text classification. Furthermore, ongoing development involves training CG-CNN on outputs from another CG-CNN to explore multi-layered architectures, aiming to construct more complex and descriptive features. This scalability and adaptability to various data types underscore the potential of CG-CNN to handle a wide range of applications, making it a promising architecture for tackling diverse data representation challenges.
\end{abstract}


\titlepgskip=-15pt

\maketitle
\section{Introduction}
Deep learning approach has led to great excitement and success in AI in recent years \cite{lecun2015deep,alzubaidi2021review,Zhao2019,goodfellow2016deep,ravi2017deep, fisher2023ai}. With the advances in computing power, the availability of manually labeled large data sets, and a number of incremental technical improvements, deep learning has become an important tool for machine learning involving big data 
\cite{Zhao2019,goodfellow2016deep, deep_expert, hu2023machine}. Deep Convolutional Neural Networks (CNNs), organized in series of layers of computational units, use local-to-global pyramidal architecture to extract progressively more sophisticated features in the higher layers based on the features extracted in the lower ones \cite{Zhao2019,goodfellow2016deep}. Such incrementally built-up features underlie the remarkable performance capabilities of deep CNNs.

When deep CNNs are trained on gigantic datasets to classify millions of data into thousands of classes, the features extracted by the intermediate hidden layers – as opposed to either the raw input variables or the task-specific complex features of the highest layers – come to represent efficiently the objective content of the data \cite{deep_expert,bengio2012deep, mukhopadhyay2023deep, zhu2023transfer, iman2023review}. Such objectively significant and thus inferentially powerful features can be used not only in the classification task for which they were developed, but in other similar classification tasks as well. In fact, having such features can reduce complexity of learning new pattern recognition tasks \cite{favorov2004sinbad}. Indeed, taking advantage of this in the process known as transfer learning \cite{deep_expert, yosinski2014transferable, zhu2023transfer, iman2023review, goyal2023texture, chen2023rock, unver2023novel, poyatos2023evoprunedeeptl, salehi2023study, tang2023deep} helps extract broad-purpose features to preprocess the raw input data and boost the efficiency and accuracy of special-purpose machine learning classifiers trained on smaller datasets \cite{Zhao2019,goodfellow2016deep, bengio2012deep, priya2023pt}. Transfer learning is accomplished by first training a ``base'' broad-purpose network on a big-data task and then transferring the learned features/weights to a special-purpose ``target'' network trained on new classes of a generally smaller ``target'' dataset \cite{yosinski2014transferable}. 

Learning generalizable representations is improved with data augmentation, by incorporating variations of the training set datasets using predefined transformations \cite{Ref90}. Feature invariances, which have long been known to be important for regularization \cite{ zhu2023transfer, iman2023review}, can also be promoted by such means as mutual information maximization \cite{ Ref80, favorov2004sinbad}, or by penalizing the derivative of the output with respect to the magnitude of the transformations to incorporate \emph{a priori} information about the invariances \cite{Ref17}, or by creating auxiliary tasks for unlabeled data \cite{selective16, dosovitskiy2014discriminative}.

Although supervised deep CNNs are good at extracting pluripotent inferentially powerful transferable features, they require big labeled datasets with detailed external training supervision. Also, the backpropagation of the error all the way down to early layers can be problematic as the error signal weakens (a phenomenon known as the gradient vanishing \cite{bottou}). To avoid these difficulties, in this paper we describe a self-supervised approach for learning pluripotent transferable features in a single CNN layer without reliance on feedback from higher layers and without a need for big labeled datasets. We demonstrate the use of this approach on two examples of a single CNN layer trained first on natural RGB image datasets and then on hyperspectral data images. Of course, there is a limit to sophistication of features that can be developed on raw input patterns by a single CNN layer. However, more complex and descriptive pluripotent features can be built by stacking multiple CNN layers, each layer developed in its turn by using our proposed approach on the outputs of the preceding layer(s). 

\textcolor{black}{Self-supervised learning enables models to learn rich representations from unlabeled data, particularly in scenarios where labeled data are scarce or costly to obtain. This method operates by creating a pretext task, such as predicting part of the data from other parts, to learn general features that can be beneficial for downstream tasks. The efficacy of self-supervised learning hinges significantly on the contextual information, the situational data that surrounds and adds meaning to the primary data features. This contextual guidance helps in learning features that are more useful for the arbitrary down-stream prediction tasks on the data. Creating and learning to discriminate self-supervised contextual meta-classes is an exhaustive task to be performed in one pass and transfer learning helps self-supervision. Transfer learning allows a model trained to perform well for one prediction problem (source domain) to serve as a starting point for a related but different problem (target domain). By integrating self-supervised learning with contextual guidance and transfer learning, the proposed methodology allows the large set of contextual classes to be broken into smaller source domain tasks that can be iteratively used to train features across varied tasks.}

This paper expands on our previous work \cite{spie, kursun_irc, kursun_ism} by enhancing and demonstrating the robustness and applicability of the Contextually Guided Convolutional Neural Networks (CG-CNN). The new contributions include:

\begin{itemize}
    \item The contextually guided model has been extended to handle datasets from different modalities, further showcasing its versatility as a generic feature extraction and transfer learning approach.

    \item New experiments on various datasets, including text classification with the 20 Newsgroups dataset, and the tactile VibTac-12 dataset have been added. These additions demonstrate CG-CNN’s ability to adapt to both structured and unstructured data across different modalities.

    \item We have introduced a new methodology for estimating the usefulness of CNN features, advancing the theoretical framework of our approach and offering a new metric, "transfer utility," which quantifies the effectiveness of the CG-CNN in various self-supervised learning settings.

    \item We have added a demonstration of the model's performance on the classical XOR problem. This provides deeper insights into the model’s capability to develop features with high pluripotency and transfer utility.

\end{itemize}

In Section 2, we briefly review transfer learning and neurocomputational antecedents of our unsupervised feature extraction approach. In Section 3, we present the proposed Contextually Guided CNN (CG-CNN) method and a measure of its transfer utility. We present experimental results in Section 4 and conclude the paper in Section 5. 
\section{Background on Transferable Feature Extraction}

Deep CNNs apply successive layers of convolution (Conv) operations, each of which is followed by nonlinear functions such as the sigmoidal or ReLU activation functions and/or max-pooling. These successive nonlinear transformations help the network extract gradually more nonlinear and more inferential features. Besides their extraordinary classification accuracy on very large datasets, the deep learning approaches have received attention due to the fact that the features extracted in their first layers have properties similar to those extracted by real neurons in the primary visual cortex (V1). Discovering features with these types of receptive fields are now expected to the degree that obtaining anything else causes suspicion of poorly chosen hyperparameters or a software bug \cite{yosinski2014transferable}. 

Pluripotent features developed in deep CNN layers on large datasets can be used in new classification tasks to preprocess the raw input data to boost the accuracy of the machine learning classifier \cite{deep_expert, zhu2023transfer}. That is, a base network is first trained on a “base" task (typically with a big dataset), then the learned features/weights are transferred to a second network to be utilized for learning to classify a “target" dataset \cite{yosinski2014transferable}. The learning task of the target network can be a new classification problem with different classes. The base network's pluripotent features will be most useful when the target task does not come with a large training dataset. When the target dataset is significantly smaller than the base dataset, transfer learning serves as a powerful tool for learning to generalize without overfitting.

The transferred layers/weights can be updated by the second network (starting from a good initial configuration supplied by the base network) to reach more discriminatory features for the target task; or the transferred features may be kept fixed (transferred weights can be frozen) and used as a form of preprocessing. The transfer is expected to be most advantageous when the features transferred are pluripotent ones; in other words, suitable for both base and target tasks. The target network will have a new output layer for learning the new classification task with the new class labels; this final output layer typically uses softmax to choose the class with the highest posterior probability.

Biological neural networks like cortical areas making up the sensory cortex (similar to deep CNNs) are organized in a modular and hierarchical architecture \cite{Jeff}. Column-shaped modules (referred to as columns) making up a cortical area work in parallel performing information processing that resembles a convolutional block (convolution, rectification, and pooling) of a deep CNN. Each column of a higher-level cortical area builds its more complex features using as input the features of a local set of columns in the lower-level cortical area. Thus, as we go into higher areas these features become increasingly more global and nonlinear, and thus more descriptive \cite{favorov2004sinbad,favorov2011neocortical,priya2023pt}. 

Unlike deep CNNs, cortical areas do not rely on error backpropagation for learning what features should be extracted by their neurons. Instead, cortical areas rely on some local guiding information in optimizing their feature selection. While local, such guiding information nevertheless promotes feature selection that enables insightful perception and successful behavior. The prevailing consensus in theoretical neuroscience is that such local guidance comes from the spatial and temporal context in which selected features occur \cite{favorov2004sinbad}. The reason why contextually selected features turn out to be behaviorally useful is because they are chosen for being predictably related to other such features extracted from non-overlapping sensory inputs and this means that they capture the orderly causal dependencies in the outside world origins of these features \cite{phillips1995discovery,favorov2004sinbad,sinbadorigins}.

\section{Contextually Guided Convolutional Neural Network (CG-CNN)}

\subsection{Basic Design}
\begin{figure*}[t]
    \centering
-   \includegraphics[width=\linewidth]{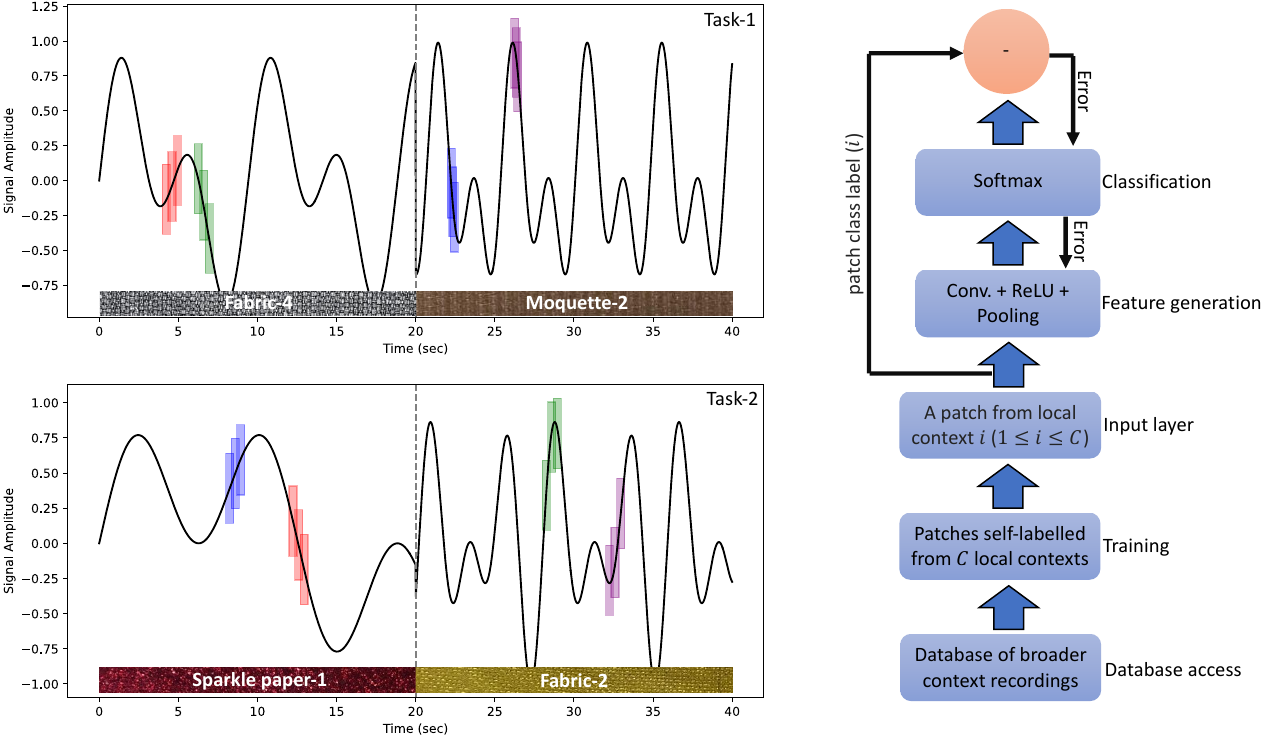}
    % Add the labels (a) and (b) below the image
    \begin{minipage}{\linewidth}
        \begin{minipage}{0.45\linewidth}
            \centering
            (a)
        \end{minipage}%
        \hfill % This ensures the two blocks are side by side
        \begin{minipage}{0.45\linewidth}
            \centering
            (b)
        \end{minipage}
    \end{minipage}
    \caption{\textbf{(a)} Class-defining contextual groups of data patches. Each patch is shown as a small rectangular box superimposed on one of the database of recordings. Neighboring patches constitute a contextual group and during network training are treated as belonging to the same class. During network training, locations of contextual groups are picked at random. Two tasks and four groups per task are shown on this photo with three patches in each ($C=4$ and $N=3$).  \textbf{(b)} CG-CNN architecture. }
    \label{fig:figure1}
\end{figure*}

In this paper we apply the cortical context-guided strategy of developing pluripotent features in individual cortical areas to individual CNN layers. To explain our approach, suppose we want to develop pluripotent features in a particular CNN layer (performing convolution + ReLU + pooling) on a given data repository. We set up a three-layer training system (Figure~\ref{fig:figure1}) as:

\begin{enumerate}
\item The Input layer, which might correspond to a 2-dimensional field of raw pixels (i.e., a 3D tensor with two axes specifying row and column and one axis for the color channels) or the 3D tensor that was outputed by the preceding CNN layer with already developed features;
\item The CNN layer (``Feature Generator''), whose features we aim to develop;
\item The Classifier layer, a set of linear units fully connected with the output units of the CNN layer, each unit (with softmax activation) representing one of the training classes in the input patterns.
\end{enumerate}

\noindent
As in standard CNNs, during this network's training the classification errors will be backpropagated and used to adjust connection weights in the Classifier layer and the CNN layer. List of the symbols used in the rest of this section are summarized in Table~\ref{tab:nomencl}.

\begin{table}[t]
    \centering
    \caption{List of Symbols}
    \begin{tabular}{|l p{5.5cm}|}
        \hline
        \textbf{Symbol}      & \textbf{Description}     \\ \hline
        $a$ & the width of the input bitmap data patches \\ \hline
        $A_{CG}(C)$ & transferable classification accuracy \\ \hline
        $b$ & the number of channels (e.g., red-green-blue) of the input bitmap data patches \\ \hline
        $C$ & the number of contextual groups \\ \hline
        $d$ & the number of feature maps (the features extracted from the $a\times a\times b$ input tensor %and fed to the Classifier) 
        \\ \hline
        $g$ & the extent of the spatial translation within contextual groups \\ \hline
        $N$ & the number of randomly chosen data patches in each contextual group \\ \hline
        $r^t$ & one-hot vector for the class label of $x^t$, where $r_c^t=1$ if and only if $x^t \in$ contextual group $c$ \\ \hline
        $s$ & the stride of the convolutions \\ \hline
        $U$ & Transfer Utility, estimates the pluripotency of the learned convolutional features \\ \hline
        $V_l$ & softmax weights of Layer-5 for contextual group $l \in \{1,2,\dots,C \}$ for the current task of discriminating $C$ groups (each softmax unit has $d$ input connections coming from $y$) \\ \hline
        $w$ & the kernel size of the convolutions \\ \hline
        $W_j$ & convolutional weights of unit $j$ in Layer-2 (each unit/feature has $w \times w\times b$ input connections) \\ \hline
        $x^t$ & data patch number $t$ used as input to CG-CNN \\ \hline
        \textbf{$\mathbf{X_E}$} & the training dataset formed by $x^t$ data patches ($1\leq t \leq C\times N$) from $C$ contextual groups with $N$ patches in each group %. This E-dataset is used for optimizing $V$ weights of Layer-5 in the E-step of the EM iterations 
        \\ \hline
        \textbf{$\mathbf{X_M}$} & the dataset formed similarly as $\mathbf{X_E}$, which is used after the E-step for estimating the goodness-of-fit, $A_{CG}$, of the current Layer-2 weights $W$. It is also used for updating the Layer-2 weights ($W_{new}$) in the M-step \\ \hline
        $y^t$ & the $d$-dimensional feature vector computed as the output of the CNN layer \\ \hline
        $y_j^t$ & the response of unit/feature $j$ in the CNN layer with $j \in \{1, 2,\dots,d\}$ \\ \hline
        $z_l^t$ & the response of output unit $l$ in the Classifier layer to the input pattern $x^t$ (estimate of the probability that $x^t$ belongs to contextual group $l$) \\ \hline
    \end{tabular}
    \label{tab:nomencl}
\end{table}

While eventually (after its training) this CNN layer might be used as a part of a deep CNN to discriminate some particular application-specific classes of input patterns, during the training period the class labels will have to be assigned to the training input patterns internally; i.e., without any outside supervision. Adopting the cortical contextual guidance strategy, we can create a training class by picking at random a set of neighboring window patches (Figure~\ref{fig:figure1}). Being close together or even overlapping, such patches will have a high chance of showing the same object and those that do will share something in common (i.e., the same context). Other randomly chosen locations in the dataset – giving us other training classes – will likely come from other objects and at those locations the neighboring window patches will have some other contexts to share. We can thus create a training dataset $\mathbf{X}=\{x^t \mid 1 \leq t \leq CN\}$ of $C\times N$ class-labeled input patterns by treating $C$ sets of $N$ neighboring window patches – each set drawn from a different randomly picked data location – as belonging to $C$ training classes, uniquely labeled from 1 to $C$. These inputs are small $a\times a\times b$ tensors, $a\times a$ patches (feature-maps) with $b$ features. We will refer to each such class of neighboring data patches as a ``contextual group.''

\vspace{3mm}

Upon a presentation of a particular input pattern $x^t$ from the training dataset $\mathbf{X}$, the response of the CNN layer is computed as:

\begin{equation}
    y_j^t = MaxPool([W_j * x^t]^+)
    \label{eq:eq1}
\end{equation}

\noindent
where $y_j^t$ is the response of output unit $j$ in the CNN layer with $d$ units (i.e., $y_j$ is CNN's feature $j$, where $1 \leq j \leq d$),  $W_j$ is the input connection weights of that unit (each unit has $w \times w\times b$ input connections), symbol * denotes convolution operation, and $[.]^+=\max\{.,0\}$ denotes the ReLU operation. Next, the response of the Classifier layer is computed by the softmax operation as:

\begin{equation}
    z_l^t = \frac{\exp{(V_l \boldsymbol{\cdot} y^t)}}{\sum_{c=1}^C \exp{(V_c \boldsymbol{\cdot} y^t)}}
    \label{eq:eq2}
\end{equation}

\noindent
where $z_l^t$ is the response of output unit $l$ in the Classifier layer (expressing the probability of this input pattern $x^t$ belonging to class $l$), $y^t = [y_j^t]_{j=1}^d$ is the $d$-dimensional feature vector computed as the output of the CNN layer, and $V_l$ is the vector of connection weights of that unit from all the $d$ units of the CNN layer. 

During training, connection weights $W$ and $V$ are adjusted by error backpropagation so as to maximize the log-likelihood (whose negative is the loss function):

\begin{equation}
    \mathcal{L}\big(V, W \big| \mathbf{X}\big) = \sum_{t=1}^{CN} {\sum_{c=1}^C r_c^t \log z_c^t}
    \label{eq:eq3}
\end{equation}

\noindent
where $r_c^t \in {0,1}$ indicates whether input pattern $x^t$ belongs to class $c$.

Table \ref{tab:nomencl}, listed in alphabetical order, provides the nomenclature of symbols used in our description of the CG-CNN algorithm.

\subsection{Iterative Training Algorithm}
If we want to develop pluripotent features in the CNN layer that will capture underlying contextual regularities in the data collections, it might be necessary to create tens of thousands of contextual groups for the network's training \cite{dosovitskiy2014discriminative,selective16}. We can avoid the complexity of training the system simultaneously on so many classes by using an alternative approach, in which training is performed over multiple iterations, with each iteration using a different small sample of contextual groups as training classes \cite{metalearn}. That is, in each iteration a new small (e.g., $C=100$) number of contextual groups is drawn from the database and the system is trained to discriminate them. Once this training is finished, a new small number of contextual groups is drawn and training continues in the next iteration on these new classes \underline{without} resetting the already developed CNN connection weights. 

For such iterative training of the CG-CNN system, we use an expectation-maximization (EM) algorithm \cite{EM,MLbook}. The EM iterations alternate between performing an expectation (\textbf{E}) step and a maximization (\textbf{M}) step. At each EM iteration, we create a new training dataset \mbox{$\mathbf{X}=\{x^t \mid 1 \leq t \leq CN\}$} of $C\times N$ \underline{self-class-labeled} input patterns and randomly partition it into two subsets; one subset $\mathbf{X_E}$ to be used in the E-step, the other subset $\mathbf{X_M}$ to be used in the M-step. Next, we perform the E-step, which involves keeping $W$ connection weights from the previous EM iteration ($W_{old}$), while training $V$ connections of the Classifier layer on the newly created $\mathbf{X_E}$ subset so as to maximize its log-likelihood $\mathcal{L}$ (Eq.~\ref{eq:eq3}):

\begin{equation}
    \text{E-step:  } V_{new}=\argmax_V \mathcal{L}\big(V, W_{old} \big| \mathbf{X_E}\big)
    \label{eq:eq4}
\end{equation}

Next, we perform the M-step, which involves holding the newly optimized $V$ connection weights fixed, while updating $W$ connections of the CNN layer on the $\mathbf{X_M}$ subset so as to maximize log-likelihood $\mathcal{L}$ one more time:

\begin{equation}
    \text{M-step:  } W_{new}=\argmax_W \mathcal{L}\big(V_{new}, W \big| \mathbf{X_M}\big)
    \label{eq:eq5}
\end{equation}

Overall, EM training iterations help CG-CNN take advantage of transfer learning and make it possible to learn pluripotent features using a small number of classes in the Classifier layer (each softmax unit in this layer represents one class). By continuing to update the CNN layer weights $W$, while the contextual groups to be discriminated by the Classifier keep changing with every EM iteration, CG-CNN spreads the potentially high number of contextual groups (classes) needed for learning data archive contextual regularities into multiple iterations \cite{metalearn}. The proposed EM algorithm for training CG-CNN achieves an efficient approach to learning the regularities that define contextual classes, which otherwise would theoretically require a $C$ value in orders of tens of thousands \cite{dosovitskiy2014discriminative}.

\begin{comment}
Note that a task-specific network optimizes its $V$ and $W$ weights by maximizing the log-likelihood for discriminating the $C$ classes given in that particular task as:
\begin{equation}
    V_{specific},W_{specific} = \argmax_{V,W} \mathcal{L}\big(V, W \big| \mathbf{X}\big)
    \label{eq:eq6}
\end{equation}
However, the $W^*$ weights found by a task-specific training will not be the most effective for discriminating a different set of $C$ classes (in a different task). 
\end{comment}

To monitor the progress of CG-CNN training across EM iterations – so as to be able to decide when to stop it – we can at each EM iteration compute the network's current classification accuracy. Since we are interested in transferability of the CNN-layer features, such accuracy evaluation should be performed after the E-step, when Classifier-layer connections $V$ have been optimized on the current iteration's task (using the $\mathbf{X_E}$ subset of input patterns), but before optimization of CNN-layer connections $W$ (which were transferred from the previous EM iteration). Furthermore, classification accuracy should be tested on the new, $\mathbf{X_M}$, subset of input patterns. Such classification accuracy can be expressed as the fraction of correctly classified test ($\mathbf{X_M}$) input patterns. We will refer to such classification accuracy of CG-CNNs with task-specific Classifier weights $V$ but transferred CNN feature weights $W$ as ``transferable classification accuracy'' and use it in Section 4 as an indicator of the usefulness of context-guided CNN features on a new task:

% FloatBarrier to prevent floats from moving past this point
\FloatBarrier

\begin{equation}
    A = \frac{1}{|\mathbf{X_M}|}\sum_{t=1}^{|\mathbf{X_M}|}\big[\underset{c}{\argmax{r_c^t}}=\underset{c}{\argmax{z_c^t}} \big]
\label{eq:A}
\end{equation}

\noindent 
where the argmax operators return the indices of the expected and predicted classes of $x^t$, respectively; and $[i=j]$ is the Kronecker delta function (expressed using the Iverson bracket notation) used to compare the expected and predicted classes. 

As will be detailed further in Section 4, no particular CNN architecture is required for applying the CG-CNN training procedures. In Algorithm~\ref{alg:CG-CNN}, we formulate the CG-CNN algorithm using a generic architecture (that somewhat resembles AlexNet because GoogLeNet, for example, does not use ReLU but it uses another layer called BatchNorm). Regardless of the particulars of the chosen architecture, CG-CNN accepts a small $a\times a \times b$ tensor as input. Although CG-CNN can be applied repeatedly to extract higher level features on top of the features extracted in the previous layer as mentioned at the beginning of this section, in this paper, focusing on CG-CNN's first application to data directly, $b$ simply denotes the number of color bands, i.e. $b=3$ for RGB image data, and $a$ denotes the width of the resemblance patches that form the contextual groups. The kernel size of the convolutions and the stride are denoted by $w$ and $s$, respectively. In CNNs, pooling operations, e.g. MaxPool, are used to subsample (that leads to the pyramidal architecture) the feature maps. A 75\% reduction is typical, which is achieved via pooling with a kernel size of 3 and stride of 2, which gives us $a=w+2s$. For example, if $w=11$ and $s=4$ for the convolutions (as in AlexNet), then $a=19$. That is, CG-CNN's Feature Generator (the CNN layer) learns to extract $d$ features (e.g., $d=64$) that most contextually and pluripotently represent any given $a\times a$ data image patch. Note that at this level CG-CNN is not trying to solve an actual classification problem and is only learning a powerful local representation; only a pyramidal combination of these powerful local features can be used to describe a data big enough to capture real-world object class.

\begin{algorithm}[t]
    \caption{The proposed CG-CNN method for learning broad-purpose transferable features}
    \label{alg:CG-CNN}
    \begin{algorithmic}[1]
        \State $\operatorname{CG-CNN}$ = [
        \State \quad \textit{// the Input layer}
        \State \quad Layer-1: InputLayer(input\_size = $a \times a \times b$)
        
        \State \quad \textit{// the CNN layer (Feature Generator)}
        \State \quad Layer-2: ConvLayer(kernel\_size = $w \times w \times b$, output\_size = $d$, stride = $s$)
        \State \quad Layer-3: ReLULayer
        \State \quad Layer-4: $y$ = MaxPool(kernel\_size = $3 \times 3$)
        
        \State \quad \textit{// the Classifier layer (Discriminator)}
        \State \quad Layer-5: $z$ = Softmax(output\_size = $C$) 
        \State ]
        
        \State Randomly initialize Layer-2 weights $W$
        
        \Repeat \textit{// Start a new EM iteration}
        
            \State \textit{// Create New Task}
            \State \textit{// Populate a new dataset, $\mathbf{X}$ with $C$ classes and $N$ instances per class}
            
            \For{$c$ = 1 to $C$}
                \State Pick a random $a \times a$ window as the seed
                
                \State \textit{Spatial Contextual Guidance:}
                \State Randomly slide the seed $\pm g$ pixels to create $N$ instances of class-$c$.
                
                \State \textit{Color-based Contextual Guidance:}
                \State Random 10\% color-jitter to perturb brightness, contrast, saturation, and hue. 50\% are converted to grayscale.
            \EndFor
            
            \State Split $\mathbf{X}$ into the E-dataset and M-dataset
            
            \State \textit{// E-step:}
            \State Set learning rate of Layer-2 to 0
            \State Randomly initialize Layer-5
            \State Train Layer-5 using E-dataset to get $V_{new}$
            
            \State Compute Accuracy, $A$, using $V_{new}$ and $W$ on M-dataset
            
            \State \textit{// M-step:}
            \State Set learning rate of Layer-5 to 0
            \State Restore learning rate for $W$ of Layer-2
            \State Fine-tune $W$ by using the M-dataset
            
        \Until $A$ converges
    \end{algorithmic}
\end{algorithm}

\vspace{1mm}

\subsection{Pluripotency Estimation of CNN Features}
EM training of CG-CNN aims to promote pluripotency of features learned by the CNN layer; i.e., their applicability to new classification tasks. Ideally, pluripotency of a given set of learned features would be measured by applying them to a comprehensive repertoire of potential classification tasks and comparing their performance with that of: (1) naïve CNN-Classifiers, whose CNN-layer connection weights are randomly assigned (once randomly assigned for a given task, these $W$ weights are never updated as in extreme learning machines \cite{wang2022review}; \cite{xavier} present a state-of-the-art weight initialization method); and (2) task-specific CNN-Classifiers, whose CNN-layer connections are specifically trained on each tested task. The more pluripotent the CG features, the greater their classification performance compared to that of random features and the closer they come to the performance of task-specific features. Such a comprehensive comparison, however, is not practically possible. Instead, we can resort to estimating pluripotency on a more limited assortment of tasks, such as for example discriminating among newly created contextual groups (as was done in EM training iterations). Regardless of the $C$-parameter used in the CG-CNN training tasks, these test tasks will vary in their selection of contextual groups as well as the number ($C$) of groups. 

\begin{figure}[ht]
    \centering
    \includegraphics[width=\linewidth]{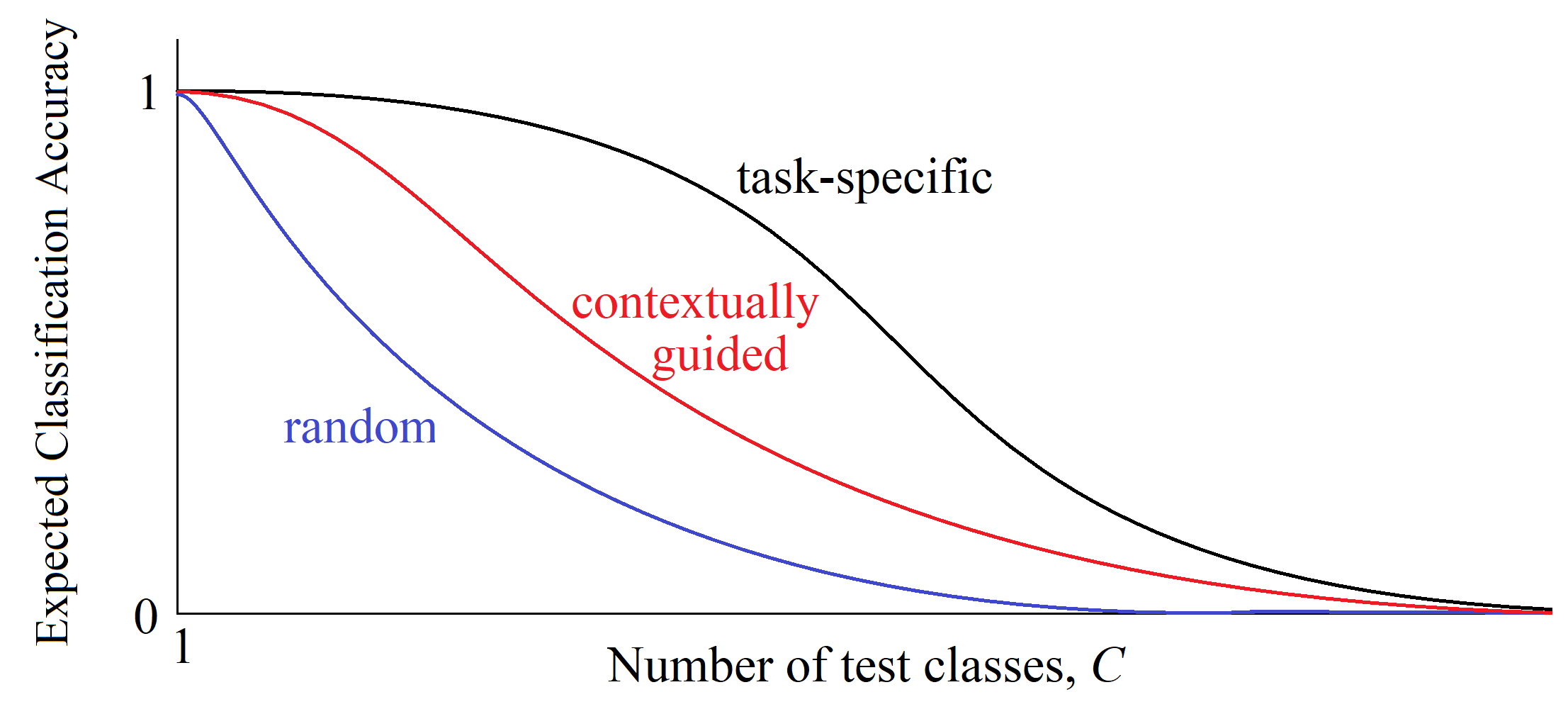}
    \caption{Transfer Utility of CG-CNN features is based on the area under the curve of the test accuracy $A_{CG}$ as a function of the number of test classes $C$. Accuracies obtained using the random and task-specific CNN features, $A_{random}(C)$ and $A_{specific}(C)$ are also shown as they are used in Eq.~\ref{eq:eq7} to quantify the Transfer Utility, $U$. The expectation of the test accuracies is computed over a number of tasks generated for each value of $C$.}
    \label{fig:figure2}
\end{figure}

The expected outcome is graphically illustrated in Figure~\ref{fig:figure2}, plotting expected classification accuracy of CNN-Classifiers with random, task-specific, and CG features as a function of the number of test classes. When the testing tasks have only one class, all three classifiers will have perfect accuracy. With increasing number of classes in a task, classification accuracy of random-feature classifiers will decline most rapidly, while that of task-specific classifiers will decline most slowly, although both will eventually converge to zero. CG-feature classifiers will be in-between. According to this plot, the benefit of using CG features is reflected in the area gained by the CG-feature classifiers in the plot over the baseline established by random-feature classifiers. Normalizing this area by the area gained by task-specific classifiers over the baseline, we get a measure of ``Transfer Utility'' of CG features:    

\begin{equation}
    U = \frac{\sum_{C=1}^{\infty}{\mathbb{E}\big[A_{CG}{(C)} \big]}-\sum_{C=1}^{\infty}{\mathbb{E}\big[A_{random}{(C)} \big]}}{\sum_{C=1}^{\infty}{\mathbb{E}\big[A_{specific}{(C)} \big]}-\sum_{C=1}^{\infty}{\mathbb{E}\big[A_{random}{(C)} \big]}}
    \label{eq:eq7}
\end{equation}

\noindent
where $A_{random}(C)$, $A_{specific}(C)$, and $A_{CG}(C)$ are classification accuracies of CNN-Classifiers with random, task-specific, and CG features, respectively, on tasks involving discrimination of $C$ contextual groups (Eq.~\ref{eq:A}).

\subsection{Sources of Contextual Guidance}
While in our presentation of CG-CNN so far we have explained the use of contextual guidance on an example of spatial proximity, using neighboring data patches to define training/testing classes, any other kind of contextual relations can also be used as a source of guidance in developing CNN-layer features. Temporal context is one such rich source. Frequency domain context is another source, most obviously in speech recognition, while in Section 4 we exploit it in a form of hyperspectral imaging. More generally, any natural phenomena in which a core of indigenous causally influential factors are reflected redundantly in multivariable raw sensor data will have contextual regularities, which might be possible to use to guide feature learning in the CNN layer.

With regard to spatial-proximity-based contextual grouping, it is different from data augmentation used in deep learning \cite{dosovitskiy2014discriminative}. Data augmentation does shift input data patches a few pixels in each direction to create more examples of a known object category (such as a car or an animal); however, for a contextual group, we do not have such object categories to guide the placement of data patches and we take patches over a much larger range pixels from the center of the contextual group. Training CG-CNN using short shifts (similar to data-augmentation) does not lead to tuning to V1-like features because other/suboptimal features can also easily cluster heavily overlapping data patches. 

Another source of contextual information that we utilize in this paper for extracting features from color data images is based on multiple pixel-color representations (specifically, RGB and grayscale). Instead of using a feature-engineering approach that learns to extract color features and grayscale features separately, as in \cite{alzubaidi2021review}, we use a data-engineering approach by extending the contextual group formation to the color and gray versions of the data image windows: For every contextual group, some of the RGB data image patches are converted to grayscale. This helps our network develop both gray and color features as needed for maximal transfer utility: if the training is performed only on gray data images, even though the neurons might have access to separate RGB color channels, whose weights are randomly initialized and the visualization of feature weights initially looks colorful, they all gradually move towards gray features. Using no grayscale data images leads to all-color features automatically. For our experiments in Section 4, the probability for the random grayscale transformation was set to 0.5. That is, we converted 50\% of the data image patches in each contextual group from color to gray, which led to emergence of gray-level features in addition to color ones.
 \section{Experimental Results}

\begin{figure*}[t]
    \centering
    \includegraphics[width=\linewidth]{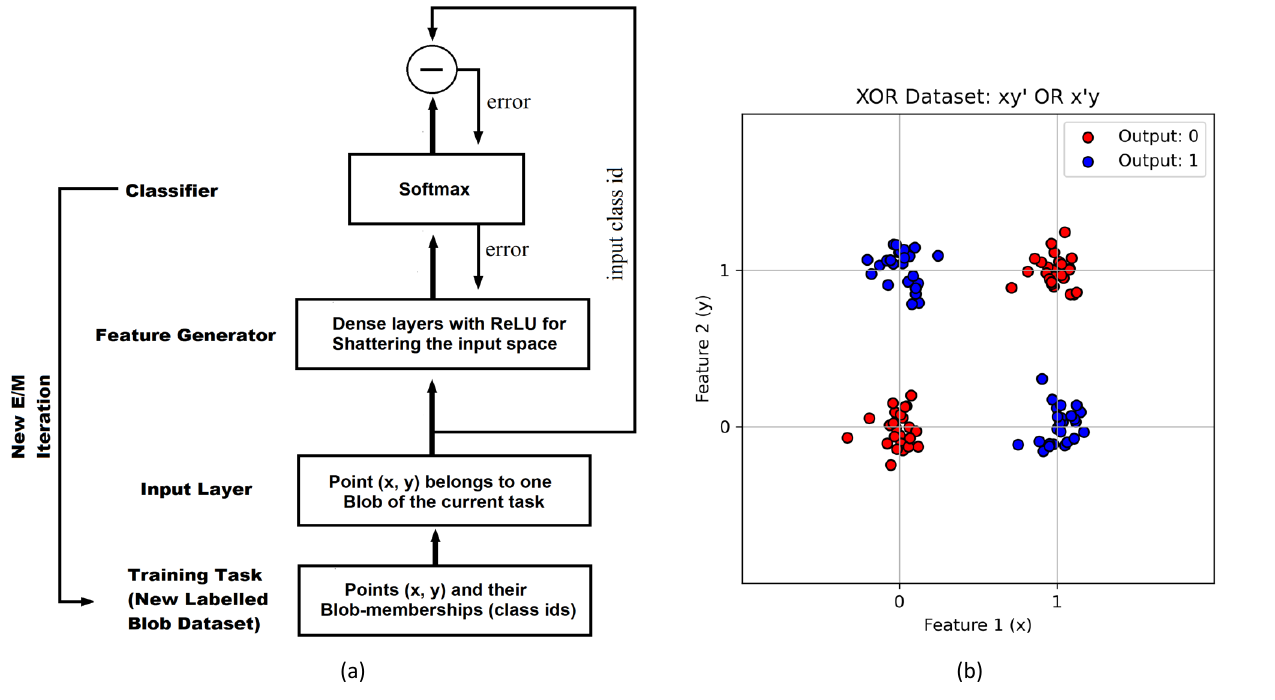}
    \caption{(a) Illustration of the CG-CNN architecture applied to the input space of the XOR dataset; (b) the XOR dataset represented using Gaussian blobs with a standard deviation of 0.1.}
        \label{fig:fig3-XOR-combined}
\end{figure*}

\subsection{Demonstration on the XOR problem} 
The XOR problem stands as a fundamental example to demonstrate the importance of extracting nonlinear features and the effectiveness of multilayer networks. In addressing this problem, our method showcases its robust representational capacity, enhancing the variety of nonlinear mappings that can be captured. This capacity is akin to the principles seen in Radial Basis Function (RBF) networks or the concept of pluripotency (high Rademacher complexity) \cite{favorov2011neocortical}, emphasizing the preservation of nuanced neighborhood structures within the data. We utilize synthetic data to illustrate the efficacy of our approach, underscoring that its success is not reliant on convolutional processes or image data, which are common in CNN applications. Instead, this demonstration highlights how our methodology can discern patterns and learn relationships in data that are not inherently image-based, showcasing its adaptability and broader applicability beyond the conventional scope of image-focused neural networks. As outlined in Figure~\ref{fig:fig3-XOR-combined}.a, this subsection presents the CG-CNN approach to solving the XOR problem through transfer learning, beginning with self-supervision on a task involving classification of random blobs, followed by fine-tuning for the XOR classification problem. 

The initial phase of our experiment involved training a neural network in a self-supervised manner on a dataset comprising randomly generated blobs (shown in Figure~\ref{fig:fig3-XOR-combined}.b). Each blob represented a distinct class, and the goal was to classify points into their respective blobs. This task was designed to encourage the model to learn rich feature representations that capture the underlying structure of the data space.

Our model architecture for this phase included two hidden layers with 20 neurons each, incorporating sigmoid activation functions. The output layer for this multi-class self-supervised task utilized a softmax activation function. Even though the number of output neurons (the number of classes in a task) changes, the hidden layers gradually learn features that can lead to better starting point than random initialization in order to obtain higher average classification accuracy (i.e. Transfer Utility) with shorter training on limited training data. We also integrated Dropout layers with a 0.1 dropout rate subsequent to each hidden layer in our model for regularization and to prevent over-reliance on particular features, thereby promoting generalization for all the models (even more so for the task-specific ones). By randomly deactivating a subset of neurons during training, Dropout ensures that different neurons are activated across iterations, reducing the likelihood of the model entirely losing previously learned information (catastrophic forgetting) from the previous self-supervised tasks. This approach helps preserve valuable features learned during the initial phase of training when the model is subsequently adapted to new tasks.

\begin{figure}[t]
    \centering
    \includegraphics[width=\linewidth]{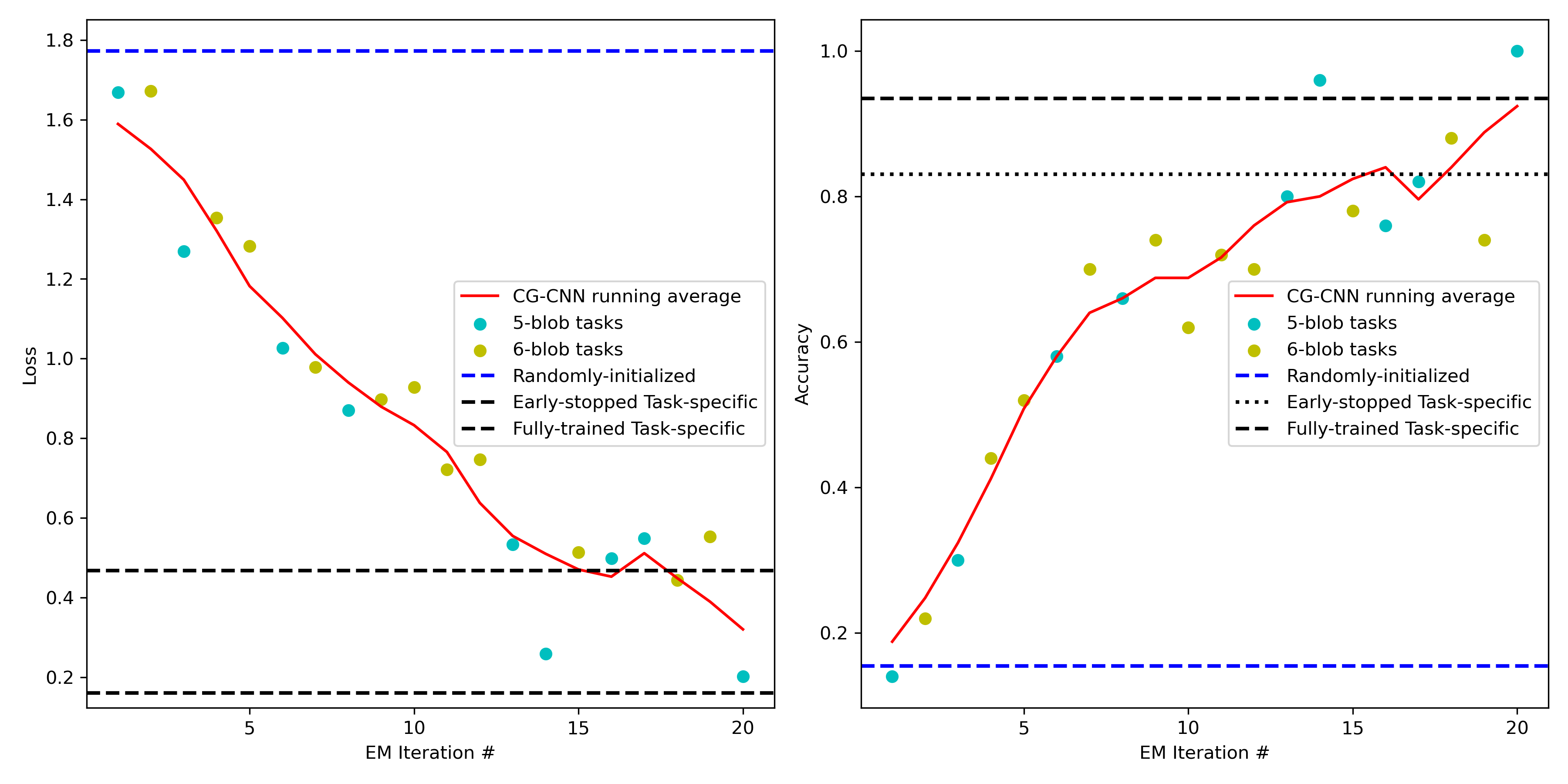}
    \vspace{-3mm}
    \caption{The transferable classification loss and accuracy are depicted in the left and right panels, respectively. As the network is exposed to more tasks, there is an evident emergence of a powerful representation, characterized by progressively lower loss and higher accuracy on self-supervised tasks that indicates a growing Transfer Utility. The smooth curve denotes the running average.}
        \label{fig:fig4-XOR-TU}
\end{figure}

\begin{figure}[t]
    \centering
    \includegraphics[width=.85\linewidth]{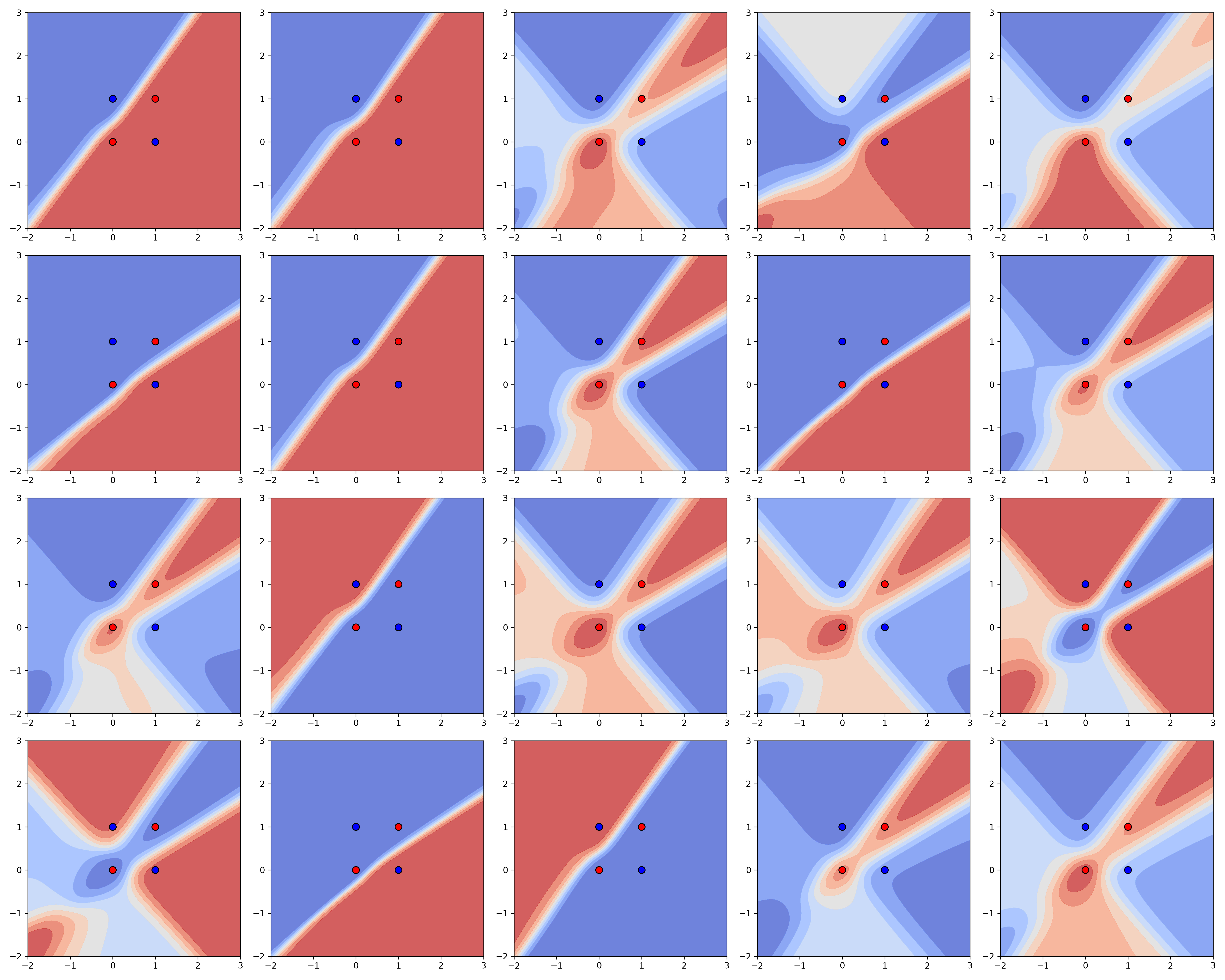}
    \caption{The task-specific features for the XOR dataset are tailored exclusively for solving the XOR problem and are not general-purpose. Unlike self-supervised learning methods, these features do not aim to develop versatile attributes applicable across a broader range of tasks.}
    \label{fig:fig5-taskspecific}
\end{figure}

\begin{figure}[t]
    \centering
    \includegraphics[width=.85\linewidth]{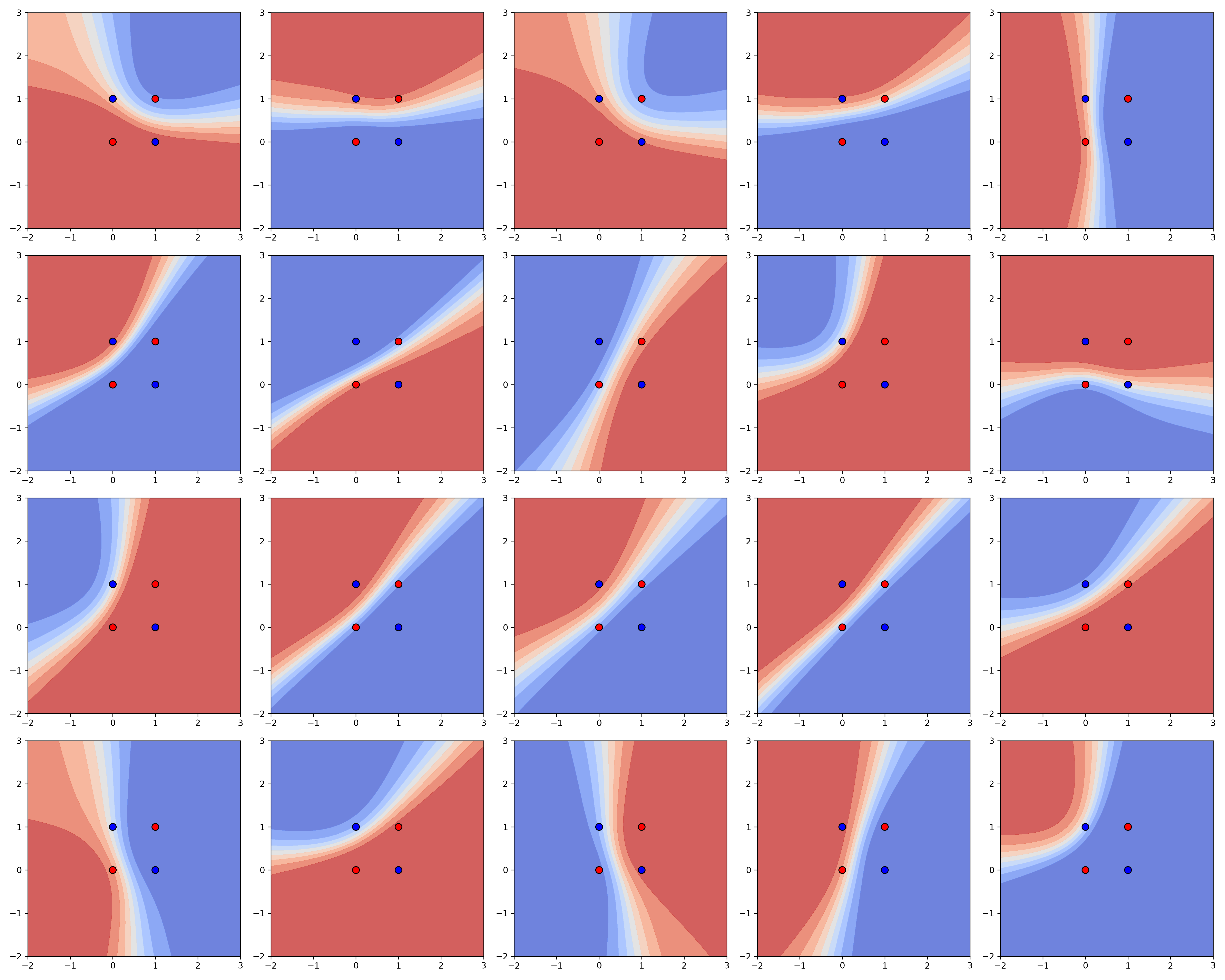}
    \caption{CG-CNN, trained on 20 diverse tasks (excluding the XOR task), learns to distinguish any patch of space from others and effectively partitions the input space more diversely. The features have the capacity to easily learn the XOR problem as well.}
    \label{fig:fig6-CGCNN}
\end{figure}

Upon completing the self-supervised learning phase, Figure~\ref{fig:fig4-XOR-TU} presents a sample run, the model develops nonlinear features of the input space that are capable of distinguishing between various blob regions (as shown and compared in Figures~\ref{fig:fig5-taskspecific} and \ref{fig:fig6-CGCNN}). We leveraged these learned features for the XOR problem (Figure~\ref{fig:fig3-XOR-combined}) as a canonical example of a non-linearly separable dataset. To adapt the model for the XOR problem, we modified the output layer to consist of a single neuron with a sigmoid activation function. Transferring the previously learned features, improved the model's ability to learn the XOR pattern. Our comparisons of the accuracy and loss of the CG-CNN based model with the random model and the task-specific model exemplifies the Figure~\ref{fig:figure2}. In Table~\ref{tab:XOR-results}, we present the average and standard deviation of accuracy and loss for 30 runs with 100 tasks and 4 to 7 blobs in each task.

\subsection{Demonstration on Natural Images} 
To demonstrate the feasibility of CG-CNN in developing pluripotent features using a limited number of images without any class-labels, we employed images from the Caltech-101 dataset \cite{bansal2023transfer}. Specifically, we utilized images from the face class to emphasize that the proposed algorithm operates without external supervision (having different classes such as human faces versus animals etc) for feature discrimination. This dataset comprised 435 color images with sizes around $400\times600$ pixels. We used half of these images to train CG-CNN and develop its features, and the other half to evaluate the pluripotency of these features. Adapting the description given in Figure~\ref{fig:figure1}, we applied the CG-CNN self-training by extracting smaller patches (e.g., 19×19 pixels) from these images as seeds and then creating a group from each sees by extracting nearby patches (shifting) and applying data augmentation techniques such as color-jitter and RGB-to-grayscale transformations. Each group was distinctively represented by a different class label, developed without relying on external supervision. The classes generated vary with each EM iteration, gradually enhancing the model's ability to find discriminatory features. Since our CG-CNN algorithm can be used with any CNN architecture, we applied it to AlexNet, ResNet, and GoogLeNet architectures. In its first convolutional block, AlexNet performs Conv+ReLU+MaxPool. This first block has $d = 64$ features, with a kernel size of $11\times11$ (i.e., $w = 11$) and a stride $s = 4$ pixels. ResNet performs Conv+BatchNorm+ReLU+MaxPool in its first block, with $d = 64$ features, kernel size of $7\times7$, and stride $s = 2$ pixels. GoogLeNet in its first block also has $d = 64$ features, $7\times7$ kernels, and stride $s = 2$. However, GoogLeNet performs Conv+BatchNorm+MaxPool. All three architectures use MaxPool with a kernel size of $3\times3$ and a stride $s = 2$. Therefore, the viewing window of a MaxPool unit is 19 pixels for AlexNet and 11 pixels for ResNet and GoogLeNet. (Note that although we could enrich these architectures by adding drop-out and/or local response normalization to adjust lateral inhibition, we chose not to do such optimizations in order to show that pluripotent features can develop solely under contextual guidance.)

\begin{table}[t]
    \centering
    \caption{Performance of models on the XOR Problem. Means and standard deviations over 30 runs are reported.}
    \label{tab:XOR-results}
    \begin{tabular}{|p{2.5cm}|p{1.5cm}|p{1.5cm}|}
        \hline
        \textbf{Model} & \textbf{Accuracy} & \textbf{Loss} \\ \hline
        CG-CNN Base with Top-Layer Trained Only & 0.9793 $\pm$ 0.0555 & 0.3678 $\pm$ 0.0852 \\ \hline
        Randomly-Initialized Base with Top-Layer Trained Only & 0.4987 $\pm$ 0.0503 & 0.6952 $\pm$ 0.0020 \\ \hline
        Task-Specific (Early-Stopped) & 0.6447 $\pm$ 0.0856 & 0.6734 $\pm$ 0.0298 \\ \hline
        Task-Specific (Highly Trained) & 0.9993 $\pm$ 0.0036 & 0.0011 $\pm$ 0.0028 \\ \hline
    \end{tabular}
\end{table}

\begin{figure}[t]
    \centering
    \includegraphics[width=\linewidth]{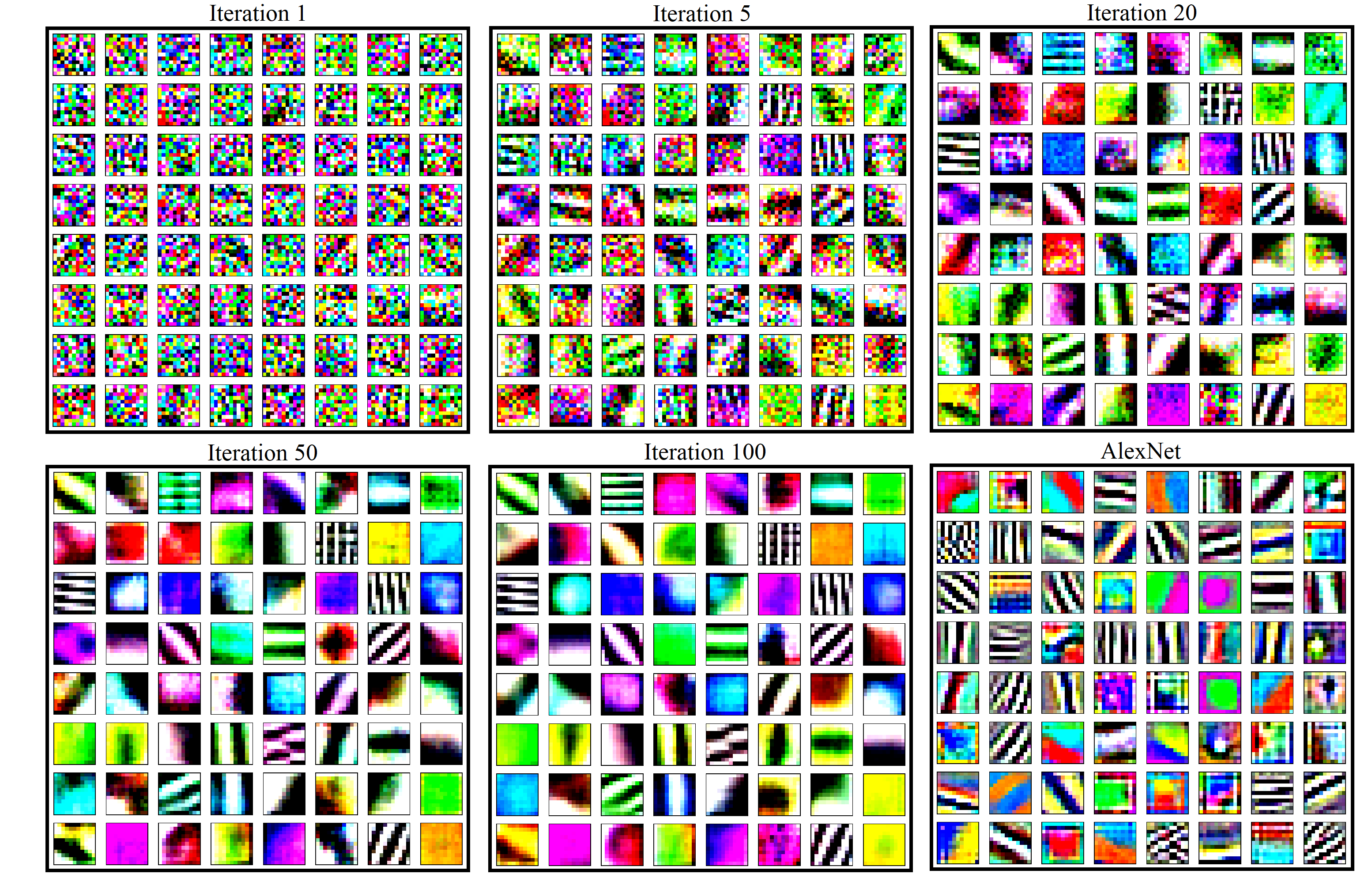}
    \caption{Visualizations of the $11\times11$ weights of the 64 features in the CNN layer of CG-CNN after 1, 5, 20, 50, and 100 EM iterations. Also shown are the weights of the 64 features in the first layer of AlexNet. While even after 20 EM iterations the features are still quite crude, the features at iterations 50 and 100 are sharp and almost identical and resemble AlexNet features.}
    \label{fig:figure5}
\end{figure}

We used a moderate number of contextual groups ($C = 100$) for the CG-CNN training. For selecting data patches for each contextual group, the parameter $g$ – used in Algorithm 1 to slide the seed window for spatial contextual guidance – was set to $g = 25$ pixels. Thus, each contextual group had $(2\times25 + 1)^2 = 2601$ distinct patch positions. We also used color jitter and color-to-gray conversion to enrich contextual groups (see Section 3.4). 

\begin{figure}[t]
    \centering
    \includegraphics[width=\linewidth]{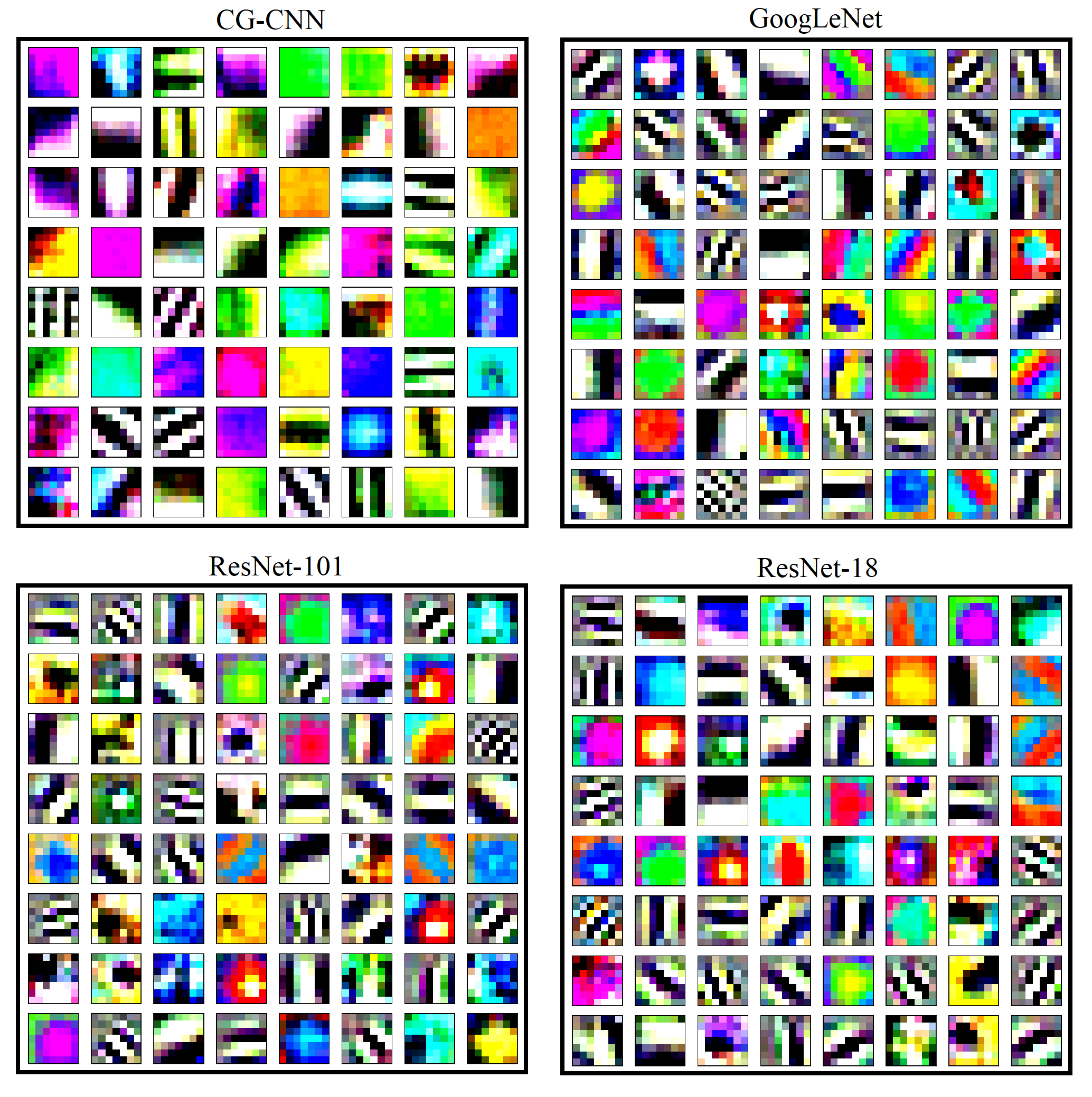}
    \caption{Visualizations of the $7\times7$  weights of the 64 features in the CNN layer of CG-CNN after 100 EM iterations, as well as 64 features in the first layer of GoogLeNet, ResNet-101, and ResNet-18.}
    \label{fig:figure6}
\end{figure}

We used PyTorch open source machine learning framework \cite{raschka2022machine} to implement CG-CNN. Experiments were performed on a workstation with Intel i7-9700K 3.6GHz CPU with 32 GB RAM and NVIDIA GeForce RTX 2080 GPU with 8GB GDDR6 memory. In each EM training iteration, we used 10 epochs for the E-step and 10 epochs for the M-step. On the workstation used for the experiments, for $C = 100$, each EM iteration takes about two minutes. CG-CNN takes around 100 minutes to converge in about 50 iterations. Both the SGD (stochastic gradient descent) and Adam \cite{adam} optimizers can reduce time. Adam helps cut down the runtime by reducing the number of epochs down to one epoch with minibatch updates. Increasing the number of EM iterations was more helpful than increasing the number of epochs in one iteration. With these improvements, 50 EM iterations took about 10 minutes, during which the Transferable Classification Accuracy, as formulated in Equation 6, initially starts around 30\% and displays an upward trend throughout the training, specifically it rises quickly in the first few EM iterations and then slowly converges to a stable level around 60\%. With each EM iteration, the network’s features become progressively more defined and more resembling visual cortical features (gratings, Gabor-like features, and color blobs) as well as features extracted in the early layers of deep learning architectures AlexNet, GoogLeNet, and ResNet (see Figures~\ref{fig:figure5} and \ref{fig:figure6}).

\begin{figure*}[t]
    \centering
    \includegraphics[width=0.45\linewidth]{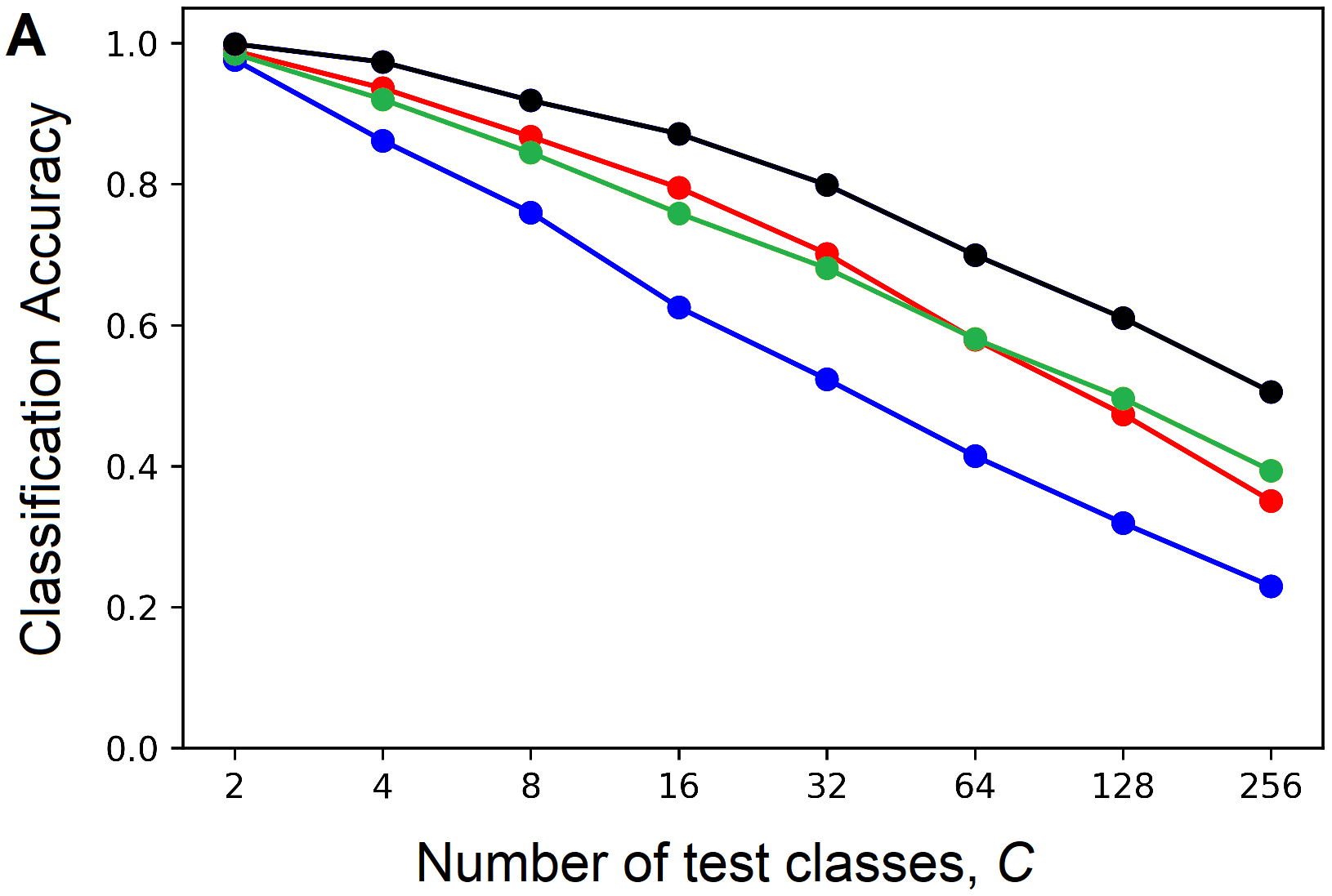}\hspace{5mm}
    \includegraphics[width=.45\linewidth]{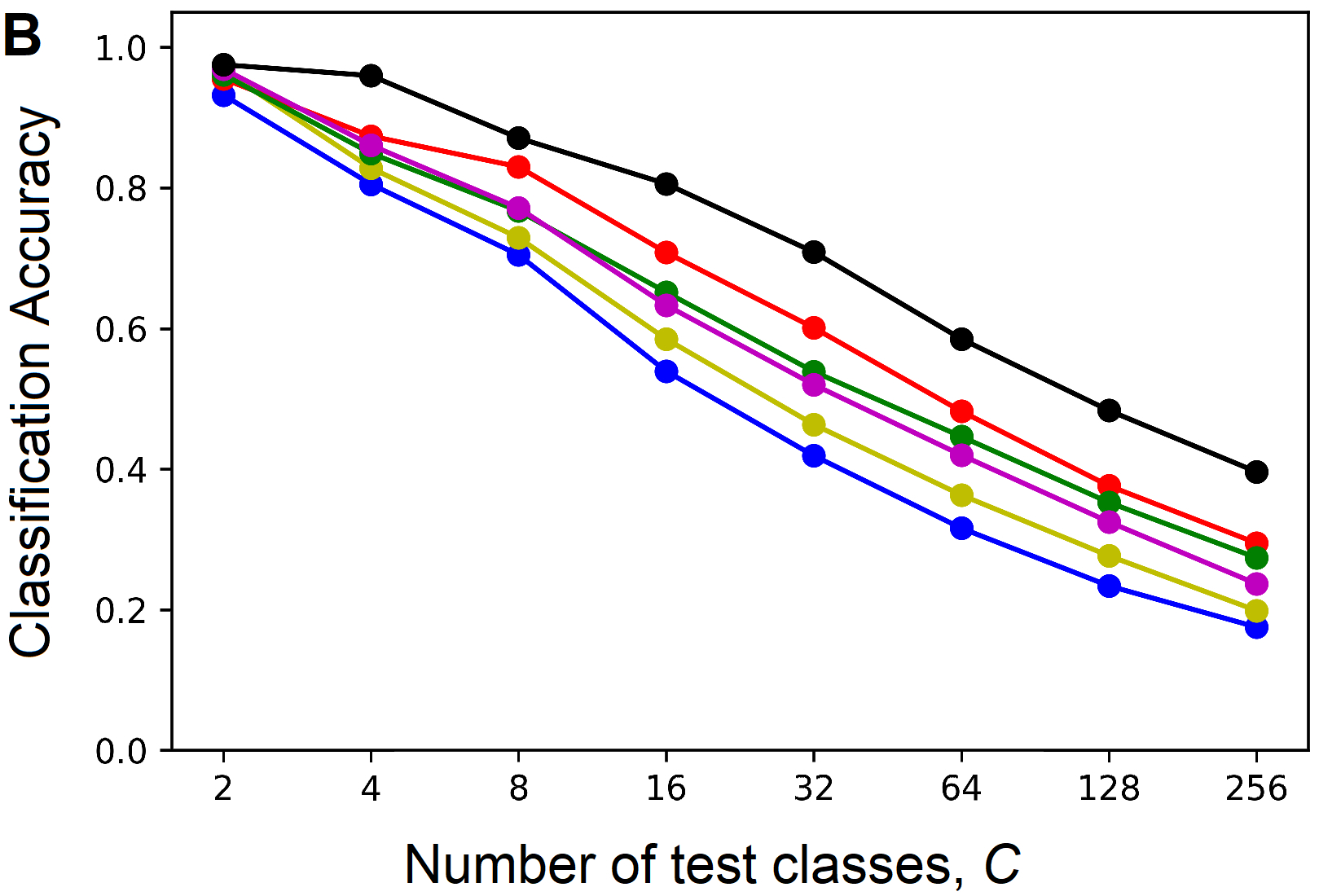}
    
    \caption{Transfer Utility of GC-CNN features, demonstrated following the format of Figure 2. \textbf{(A)} $11\times11$ pixel features. Average Classification Accuracies of CNN-Classifiers with task-specific, random, and CG features ($A_{specific}$, $A_{random}$, and $A_{CG}$; black, blue, and red curves, respectively) are plotted as a function of the number of contextual classes used in a classification task. For a comparison, also plotted is the average Classification Accuracy of CNN-Classifiers with Pool-1 features of the pretrained AlexNet (green curve). \textbf{(B)} $7\times7$ pixel features. Average Classification Accuracies of CNN-Classifiers with task-specific (black), random (blue), and CG (red) features are plotted as a function of the number of test classes. For a comparison, also plotted are the average Classification Accuracies of CNN-Classifiers with Pool-1 features of the pretrained GoogLeNet (green), ResNet-101 (magenta), and ResNet-18 (yellow) networks.}
    \label{fig:figure7}
\end{figure*}

To compare pluripotency of CG-CNN features to pluripotency of AlexNet, ResNet, and GoogLeNet features, we used the Transfer Utility approach described in Section 3.3 (see Figure~\ref{fig:figure2} and Eq. 7) and tested classification accuracy of CNN-Classifiers equipped with random, task-specific, and CG features, as well as pretrained AlexNet, GoogLeNet, and ResNet features, on new contextual groups/classes drawn from the test set that were not used during the training of CG-CNN. These classification accuracy estimates are plotted in Figure~\ref{fig:figure7} as a function of the number of test classes $C$ in each classification task. As the two plots in Figure~\ref{fig:figure7} show, the curves generated with CG-CNN features lay slightly more than halfway between the curves generated with random and task-specific features, indicating substantial degree of transfer utility. Most importantly, CG-CNN curves match or even exceed the curves generated with features taken from deep CNN systems, which are acknowledged – as reviewed in Sections 1 and 2.1 – to have desirable levels of transfer utility.

\subsection{Demonstration on Texture Classification} 

\begin{figure}[t]
    \centering
    \includegraphics[width=0.8\linewidth]{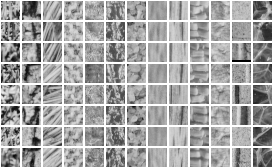}
    \caption{Brodatz texture data. Shown are 8 representative $19\times19$ pixel extracts from each of the 13 $512\times512$ pixel images in the dataset.}
    \label{fig:figure8}
\end{figure}

As an additional test of pluripotency of CG-CNN features, we applied them to a texture classification task. Texture is a key element of human visual perception and texture classification is a challenging computer vision task, utilized in applications ranging from biomedical image analysis to industrial automation and remote sensing \cite{texture_expert}. For this test, we used the Brodatz dataset \cite{brodatz} of 13 texture data images, in which each data image shows a different natural texture and is $512\times512$ pixels in size (Figure~\ref{fig:figure8}). To compare with AlexNet (which has $11\times11$ pixel features, stride $s = 4$, and therefore pooled window size of $19\times19$ pixels), we trained classifiers to discriminate textures in $19\times19$ pixel Brodatz data patches. To compare with GoogLeNet and ResNet (which have $7\times7$ pixel features, stride $s = 2$, and therefore pooled window size of $11\times11$ pixels), we trained other classifiers to discriminate textures in $11\times11$ pixel Brodatz image data patches. For either of these two window sizes, we subdivided each $512\times512$ texture image data into 256 $32\times32$ subregions and picked 128 training data patches at random positions within 128 of these subregions, and other 128 test data patches at random positions within the remaining 128 subregions. This selection process ensured that none of the training and test data patches overlapped to any degree, while sampling all the data territories.

Using the $128 \times 13 = 1664$ training data patches, we trained CNN-Classifiers equipped with either CG-CNN features (previously developed on Caltech-101 data images, as described above in Section 4.1), or AlexNet, GoogLeNet, or ResNet features. Note that these features were not updated during classifier training; i.e., they were transferred and used ``as is'' in this texture classification task. For additional benchmarking comparison, to gauge the difficulty of this texture classification task, we also applied some standard machine learning algorithms \cite{classifiers}, including Decision Trees and Random Forests, Linear and RBF SVMs, Logistic Regression, Naive Bayes, MLP (Multi-Layer Perceptron), and K-NN (K-Nearest Neighbor). 

These classifiers are straightforward to optimize without requiring many hyperparameters. For their implementation (including optimization/validation of the classifier hyperparameters), we used scikit-learn Python module for machine learning \cite{raschka2022machine}. For MLP, we used a single hidden layer with ReLU activation function (we used 64 hidden units in the layer to keep the complexity similar to that of CG-CNN). We used the default value for the regularization parameter ($C = 1$) for our SVMs and the automatic scaling for setting the RBF radius for RBF-SVM. We used $K = 1$ neighbor and the Euclidean distance metric for K-NN. For the random forest classifier, we used 100 trees as the number of estimators in the forest. 

All the classifiers were tested on the data patches from the test set, not used in classifier training. There are a total of $128 \times 13 = 1664$ test data patches. The accuracies of the classifiers are listed in Table~\ref{tab:classAcc}. According to this table, all the CNN-classifiers with transferred features had very similar texture classification accuracies, with CG features giving the best performance. All the other classifiers performed much worse, indicating non-trivial nature of this classification task. These results demonstrate the superiority of using the transfer learning approach, with transferred features taken from CG-CNN or Pool-1 of pretrained deep networks.

\subsection{Demonstration on Hyperspectral Images}

Unlike color image processing that uses a large image window with a few color channels (grayscale or RGB), Hyperspectral Image (HSI) analysis typically aims at classification of a single pixel characterized by a high number of spectral channels (bands). Typically, HSI datasets are small, and application of supervised deep learning to such small datasets can result in overlearning, not yielding pluripotent task-transferrable HSI-domain features \cite{HSI_expert}. To improve generalization, the supervised classification can benefit from unsupervised feature extraction of a small number of more complex/informative features than the raw data in the spectral channels. In this section, we demonstrate the usefulness of the features extracted by the proposed CG-CNN algorithm on two well-known HSI datasets \cite{Hyperspe47:online}: (i) The Indian Pines dataset comprises a $145\times145$ pixel image data with 224 spectral bands in the 400-2500 nm range, including 16 classes such as crops, grass, and woods; (ii) The Salinas dataset features a $512\times217$ pixel HSI image data, also with 224 spectral bands, encompassing 16 classes like vegetables, bare soils, and vineyards.

\begin{table}[t]
    \centering
    \caption{Texture classification accuracies of 12 classifiers on 13 textures taken from the Brodatz (1966) dataset. Listed are means and standard deviations of the means computed over 10 test runs.}
    \label{tab:classAcc}
    \begin{tabular}{|l|c|c|}
        \hline
        \textbf{Method} & \textbf{$11 \times 11$ field} & \textbf{$19 \times 19$ field} \\ \hline
        CG-CNN          & 63.3 $\pm$ 0.7 & 74.3 $\pm$ 0.9 \\ \hline
        AlexNet         &                 & 72.2 $\pm$ 0.7 \\ \hline
        GoogLeNet       & 62.2 $\pm$ 1.1 &               \\ \hline
        ResNet-101      & 61.6 $\pm$ 0.9 &               \\ \hline
        ResNet-18       & 61.5 $\pm$ 0.9 &               \\ \hline \hline
        RBF-SVM         & 53.6 $\pm$ 0.9 & 62.9 $\pm$ 0.9 \\ \hline
        Naive Bayes     & 39.3 $\pm$ 1.0 & 49.4 $\pm$ 0.8 \\ \hline
        Random Forest   & 34.7 $\pm$ 1.3 & 35.0 $\pm$ 1.5 \\ \hline
        MLP             & 33.6 $\pm$ 0.7 & 30.7 $\pm$ 0.6 \\ \hline
        K-NN            & 28.6 $\pm$ 0.9 & 29.2 $\pm$ 0.8 \\ \hline
        Linear-SVM      & 23.3 $\pm$ 1.2 & 28.0 $\pm$ 1.9 \\ \hline
        LR              & 22.8 $\pm$ 1.5 & 25.1 $\pm$ 0.6 \\ \hline
    \end{tabular}
\end{table}

To evaluate the quality of CG-CNN features on the hyperspectral data, the CG-CNN architecture presented in Algorithm 1 was used with the following parameters: $a = 3$ pixels, $b = 200$ bands for Indian Pines and 204 bands for Salinas (after discarding water absorption bands), $d = 30$ features, $w = 1$ pixel (i.e., each convolution uses only the bands of a single pixel), $C = 20$ contextual groups, $g = 2$ pixels for the extent of the spatial contextual guidance, and only 10 EM iterations. In training of CG-CNN, the class labels of the HSI pixels were not used; instead, local groups of pixels (controlled by the $g$ parameter) were treated as training classes, as described in Section 3.4. CG-CNN learns to represent its input HSI data patch, which is a hypercube of size $3\times3\times200$, in such a way that the data patch and its neighboring windows/positions (obtained by shifting it $g = \pm2$ pixels in each direction) can be maximally discriminable from other contextual groups centered elsewhere. Note that only a total of $(2\times2 + 1)^2 = 25$ data patches are created for each contextual group. We subsequently employed the extracted features as inputs for a Random Forest classifier. While the potential of multiple iterations of CG-CNN to produce even more descriptive and highly nonlinear features is acknowledged, this remains an avenue for future research due to its additional complexities. The primary focus of this study is to showcase the improvement attained with these features compared to the raw ones (original bands). The use of Random Forest (RF) introduces ample nonlinearity \cite{classifiers}, from which the original (raw) features can greatly benefit. However, it is observed that inputs derived from CG-CNN offer superior performance when used with RF, owing to their richer and more complex feature representation. This underscores the effectiveness of CG-CNN features in enhancing the RF classifier's performance as shown in Table~\ref{tab:rfComparison}. The results are averaged over 10 runs using a 70-30 split (excluding unlabeled pixels), with the default n\_estimators=100 hyperparameter for the Random Forest classifier. CG-CNN's ability to learn robust embeddings through self-supervision not only enhances the accuracy of the downstream classifier but also enables faster training and testing times by reducing the number of features. This simplification is particularly beneficial for real-time applications.

\begin{table}[t]
    \centering
    \caption{Comparison of RF Model Performance on the Indian Pines and Salinas Datasets.}
    \begin{tabular}{|m{3.4cm}|>{\centering\arraybackslash}m{1.25cm}|>{\centering\arraybackslash}m{1.25cm}|}
    \hline
    \textbf{Feature Type} & \textbf{Indian Pines} & \textbf{Salinas} \\ \hline
    Only Central Pixel with Original Bands ($b$ Features) & 86.3 $\pm$ 0.7  & 94.9 $\pm$ 0.1  \\ \hline
    Bands from $3\times 3$ Neighborhood Merged ($9\times b$ Features) & 91.2 $\pm$ 0.3  & 95.1 $\pm$ 0.2  \\ \hline
    CG-CNN from $3\times 3$ Neighborhood (30 Features Total) & 96.2 $\pm$ 0.6  & 97.6 $\pm$ 0.2  \\ \hline
    \end{tabular}
    \label{tab:rfComparison}
\end{table}

\vspace{5mm}

\subsection{Demonstration on VibTac-12 Vibrotactile Texture Signal Classification} 
\begin{figure*}[htbp]
    \centering
    \includegraphics[width=.85\linewidth]{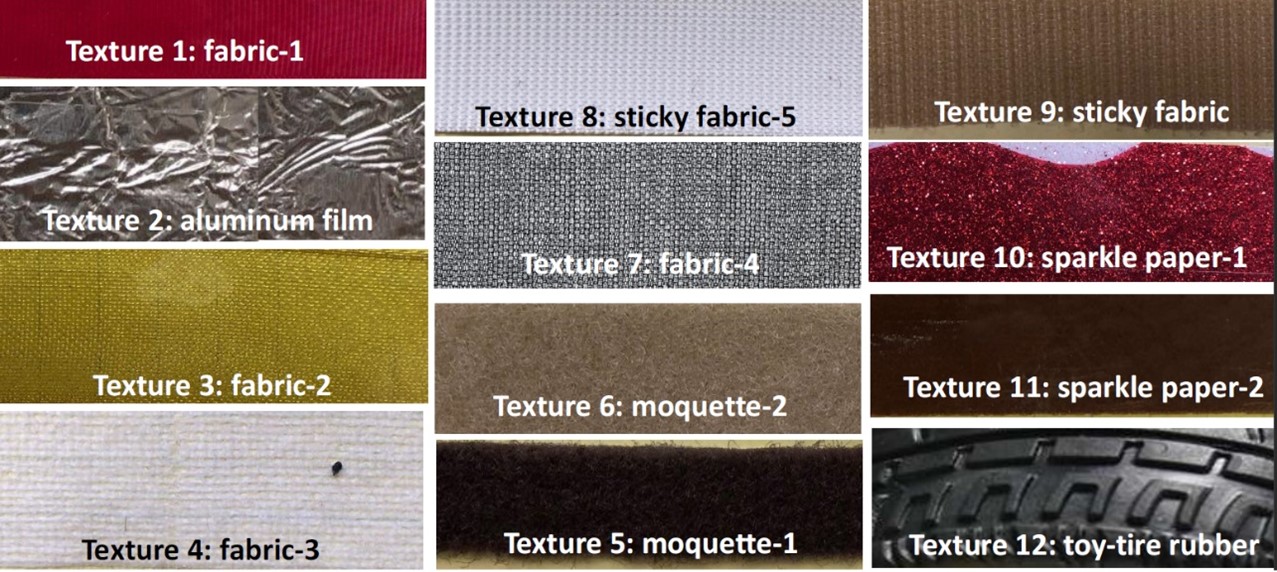}
    \caption{Images of the 12 texture classes recorded in the
VibTac-12 dataset \cite{kursun2020a}}
    \label{fig:vibTac}
\end{figure*}

In this section, we study the performance of CG-CNN on a different dataset i.e., VibTac-12 \cite{VibTac-12} which is a dataset of vibrotactile signals collected from a 3-dimensional accelerometer sensor (MMA-7660 from NXP Company \cite{accelerometer}). The signals are recorded as a probe scratches a rotating drum covered with various textures, simulating the sense of touch in humans. The diameter of the drum is \SI{7}{\centi\meter} and it rotates at a linear speed of \SI{5}{\centi\meter\per\second} which was chosen as a typical touch velocity. For each of the 12 textures, a 20-second recording is collected that corresponds to nearly five rotations of the drum. This dataset was collected and published in our previous work \cite{kursun2020a, VibTac-12} using commercial off-the-shelf embedded boards and electrical components (AVR-based embedded boards, stepper motors, etc.) as well as our own designed and 3D printed mechanical components (including the rotating drum glued with different texture strips). The collected tactile dataset has 12 texture classes and Figure~\ref{fig:vibTac} shows an exemplary subset of texture strips that were used for the experiments. Textures include sandpapers of various grits, Velcro strips with various thicknesses, aluminum foil, and rubber bands of various stickiness. 

\textcolor{black}{In this study, we employ CG-CNN as a semi-supervised approach to tackle the classification challenges posed by the VibTac-12 dataset, integrating both supervised and unsupervised learning phases using the Expectation-Maximization (EM) methodology. The experiments use labeled examples from classes 1-10 and unlabeled examples from classes 11 and 12 to evaluate the transferability and effectiveness of the learned features across different sensory signals.}

\textcolor{black}{During the semi-supervised learning phase, the model leverages EM cycles to form meta-classes based on contextual similarities among data patches while simultaneously predicting class labels (when available). The process begins by selecting a seed data patch for each meta-class and expanding these meta-classes using additional input patches through data augmentation. An additional softmax layer attempts to predict the true class labels. In the E-step, two discriminators (softmax classifiers) are trained while keeping the feature extractor frozen, and in the M-step, the features are fine-tuned while the discriminators remain static. At the end of each M-step, a new EM iteration begins, creating a new task with a refreshed set of labeled and unlabeled examples.}

\textcolor{black}{Once the pretraining is completed, fine-tuning commences on a few labeled examples from classes 11 and 12, with the features frozen to evaluate their quality for the supervised learning task. It is important to note that self-supervised training of the CG-CNN was conducted on entirely distinct sets for self-supervision, training, and testing to prevent data leakage and ensure valid experimentation. Specifically, the fine-tuning process is executed on different segments of the signals that were not used during EM training. The model's classification accuracy is then evaluated on separate segments of classes 11 and 12.}

\textcolor{black}{Additional details of the experiments are as follows. We used 2-second recordings, corresponding to 400 timesteps per example, with 500 examples per class. For pretraining, we used data from 6 to 16 seconds, leaving 0.5 to 5.5 seconds for training on classes 11 and 12, and 16.5 to 19.5 seconds for testing. The first and last 0.5 seconds were excluded to avoid noise. During pretraining, we defined 20 auxiliary classes and conducted 10 EM iterations, each consisting of 20 epochs for the E-step  (training the softmax layers while keeping the features frozen) and 20 epochs for the M-step (fine-tuning the features). For the target task (classes 11 and 12), we used 50 labeled examples per class. We employed a 1D Convolutional Neural Network (CNN) for feature extraction, with the following architecture:}

\begin{verbatim}
Conv1D(16, kernel_size=3, strides=2)
ReLU()
BatchNormalization()
MaxPooling1D(pool_size=2)
Conv1D(32, kernel_size=3, strides=2)
ReLU()
BatchNormalization()
MaxPooling1D(pool_size=2)
Conv1D(64, kernel_size=3, strides=1)
ReLU()
Dropout(0.2)
x = GlobalMaxPool1D()
supervised_output = Softmax(10)(x)
aux_output = Softmax(C)(x)
\end{verbatim}

\textcolor{black}{The model consists of three convolutional layers with 16, 32, and 64 filters, each followed by regularization layers such as batch normalization, pooling, and dropout. The network concludes with two softmax outputs: one for the supervised task and one for the auxiliary task. After pretraining with EM, we froze the learned features and fine-tuned the model on 50 labeled examples from classes 11 and 12. As described above, the pretraining, finetuning, and testing were done on separate segments to ensure no data leakage.}

As shown in Table~\ref{tab:xyz-results}, the features learned by the semi-supervised CG-CNN significantly enhance performance in the downstream classification task with labeled data.

\begin{table}[t]
\centering
\caption{\textcolor{black}{Comparisons of methods on the target domain classification on the XYZ sensor-signal data. Results are represented as an average of 10 runs and their standard deviation.}}
\label{tab:xyz-results}
\textcolor{black}{
\begin{tabular}{@{}p{5cm}p{2cm}@{}}
\toprule
\textbf{Method} & \textbf{Accuracy (\%)} \\ \midrule
Randomly Initialized CNN (trained only on the target domain with few labeled examples) & $63.1 \pm 8.3$ \\
CNN with Transfer Learning (features transferred from a labeled source domain) & $65.0 \pm 10.1$ \\
CG-CNN (trained using only contextual information from the source domain, fine-tuned on few labeled target-domain examples) & $83.9 \pm 11.4$ \\
Semi-supervised CG-CNN (trained using both labeled source data and unlabeled target-domain data, fine-tuned on few labeled target-domain examples) & $93.9 \pm 5.0$ \\ \bottomrule
\end{tabular}
}
\end{table}

\begin{itemize}
    \item \textcolor{black}{\textbf{Randomly Initialized CNN}: Trained from scratch on the target domain using random initialization with only a few labeled examples. This approach establishes a baseline performance without leveraging any pre-learned features. It has 8,097 learnable parameters, including those in the binary classification output.}
    
    \item \textcolor{black}{\textbf{CNN with Transfer Learning}: This method transfers features learned from a labeled source domain but does not utilize contextual information. It only benefits from the abundant labeled examples in the source domain, leading to some improvements in classification accuracy on the target domain. During base training, it learns 8,682 parameters. The classifier layer is then replaced with a binary output instead of the original 10-output layer. The target model has 8,097 learnable parameters, benefiting from the initialization performed during the base training.}
    
    \item \textcolor{black}{\textbf{CG-CNN with Unsupervised Feature Transfer}: Trained using only contextual information from the source domain without utilizing class labels, followed by fine-tuning on a few labeled examples from the target domain. While this method demonstrates the effectiveness of unsupervised feature learning in improving target-domain performance, it does not take advantage of the available class labels in the source domain. It learns 9,332 parameters in the base network, where global pooling reduces the output to 64 final features. The softmax layer (with $C=20$ output neurons) is then replaced with a single binary output layer, and the model fine-tunes 65 parameters of this new layer for the target task.}
    
    \item \textcolor{black}{\textbf{Semi-Supervised CG-CNN}: This approach combines both labeled source data and unlabeled target-domain data for training, followed by fine-tuning on a few labeled examples from the target domain. It achieves the highest accuracy by integrating semi-supervised learning with feature transfer. This method uses contextual information from both labeled and unlabeled examples and leverages class labels when they are available. It learns 9,982 parameters in the base network (including an extra softmax layer utilized for the supervised examples), where global pooling reduces the output to 64 final features. The top two softmax layers are then replaced with a single binary output layer, and the model fine-tunes 65 parameters of this new layer for the target task.}
\end{itemize}

\subsection{Demonstration on Word Embeddings and Text Classification}

We demonstrate the applicability of CG-CNN to word embeddings \cite{mikolov2013distributed}, which transform the discrete, symbolic nature of language into a continuous, multi-dimensional space, enabling neural networks to effectively process text. To adapt CG-CNN for word embeddings, we utilized the 20 Newsgroups dataset, a collection of documents from various newsgroup discussions. In this example, our approach not only generates contextually-guided word embeddings from scratch but also finetunes pre-existing embeddings to better suit the corpus. To optimize the dataset for neural network training, we executed the following preprocessing steps. Tokenization, conducted using a tokenizer configured to retain the top 10,000 unique tokens based on frequency, ensured the model's focus on the most relevant and prevalent terms within the corpus. Stopwords were eliminated using the list of the nltk library because these commonly occurring words hold little to no discriminative value for our analysis. Furthermore, tokens representing numerical values and those comprising fewer than three characters were removed, as they often contribute negligible contextual information and can introduce additional noise into the data. To ensure model training with meaningful context, documents shorter than 100 tokens (words) were discarded. These preprocessing measures resulted in a refined corpus of 2,961 documents with 7,811 unique tokens. The zero performance, which refers to the accuracy of a naive classifier that always predicts the majority class, was 12\%. The baseline performance when using Word2Vec involves averaging the 300-dimensional embeddings of the words in each document, followed by logistic regression (LR) as the classifier. Word2Vec+LR results in a mean accuracy of \(60.89\%\) with a standard deviation of \(4.59\%\). Similarly, when Word2Vec feature averaging is followed by a random forest (RF) classifier, it yields a mean accuracy of \(66.91\%\) with a standard deviation of \(4.15\%\). All results presented in this study are reported as averages over 10 independent runs, using different random train-test splits for each run to ensure the robustness and reliability of the findings.

For the self-supervised training of CG-CNN, we divided the entire dataset into three parts using a stratified split to preserve class priors: 50\% for the self-supervised training set where embeddings and features are learned without labels, 25\% for the supervised training set where these features are mapped to the Newsgroups classes, and the remaining 25\% for the test set to evaluate the effectiveness of this mapping on other unseen labeled examples. The CG-CNN training treats each document of the self-supervised training set as a separate meta-class, with examples from each meta-class consisting of text blobs that were randomly selected within that document. However, CG-CNN uses only $C$ classes in an EM iteration, but not all documents at the same time. In our experiments, we used $C=20$. We used text blobs of 20 tokens in length for this purpose. We explored two methods for learning feature representations: fine-tuning an existing Word2Vec embeddings and training an embedding layer from scratch.

\textbf{Fine-tuning Word2Vec:} In this approach, the aim is to fine-tune the pre-trained embeddings for capturing patterns in the data not recognized by the pre-trained Word2Vec model. We utilized the sequence of 20 300-dimensional Word2Vec embeddings as input to a neural network comprising a convolutional layer followed by a ReLU activation and a global max pooling layer. This setup reduces each blob into a single 300-dimensional vector (summarizing the context), which is then classified into \(C=20\) randomly selected meta-classes using a softmax layer. We extracted 80 blobs (sequences of 20 tokens) for each meta-class, which were evenly divided into two sets of 40 for the E and M steps of the EM iterations, respectively. After training through these EM iterations, we input a single token into the network and captured the output from the ReLU layer, recording this as the newly fine-tuned 300-dimensional embedding for that token. The specifics of the model configuration are outlined in Table~\ref{tab:code1}.

\begin{table}[t]
\centering
\caption{Configuration of a CG-CNN for fine-tuning Word2Vec’s pre-existing word embeddings.}
\label{tab:code1}
\begin{tabular}{|p{1cm}|c|p{2.5cm}|}
\hline
\textbf{Layer}            & \textbf{Parameters}                            & \textbf{Description}                        \\ \hline
\texttt{Conv 1D}           & \begin{tabular}[c]{@{}c@{}}filters=300,\\ kernel\_size=1,\\ strides=1,\\ input\_shape=(20, 300),\\ activation='relu'\end{tabular} & \begin{tabular}[l]{@{}l@{}}300 features,\\ processes tokens \\ independently,\\ no skipping of tokens,\\ ReLU activation \end{tabular} \\ \hline
\texttt{Global MaxPool 1D} & None                                          & \begin{tabular}[l]{@{}l@{}}Outputs a single \\ 300-dimensional vector \\ representing the context\end{tabular} \\ \hline
\texttt{Dense}            & \begin{tabular}[c]{@{}c@{}}units=20,\\ activation='softmax'\end{tabular} & \begin{tabular}[l]{@{}l@{}}Classifies into \\ 20 meta-classes using \\ softmax \end{tabular} \\ \hline
\end{tabular}
\end{table}

\textbf{Training a New Embedding Layer:} In contrast to the fine-tuning method, this experiment utilized a dedicated Embedding Layer instead of a Conv1D layer. This approach aims to construct embeddings from scratch by converting integer sequences (representing words) into dense vectors, which are then used to classify the context of tokens within documents. Table~\ref{tab:code2} illustrates the architecture of the model. In this model, the Embedding layer has a vocabulary size of 7811, with each word learned to be mapped to a 300-dimensional vector, and it accepts sequences of 20 tokens. GlobalMaxPooling1D is applied to reduce the spatial dimensions of the embedding output to the most significant features for each token sequence. The Dense layer then classifies these features into one of 20 meta-classes using a softmax activation function.

\begin{table}[t]
\centering
\caption{Configuration of a CG-CNN designed for learning word embeddings from scratch.}
\label{tab:code2}
\begin{tabular}{|p{1.5cm}|c|p{2cm}|}
\hline
\textbf{Layer}             & \textbf{Parameters}                            & \textbf{Description}                        \\ \hline
\texttt{Embedding}         & \begin{tabular}[c]{@{}c@{}}input\_dim=7811,\\ output\_dim=300,\\ input\_length=20\end{tabular} & \begin{tabular}[l]{@{}l@{}}Embeds 7811 \\ tokens into 300D \\ vectors,\\ sequences of 20 \\ tokens\end{tabular} \\ \hline
\texttt{Global MaxPool 1D} & None                                          & \begin{tabular}[l]{@{}l@{}}Outputs a single \\ 300D vector  \\representing the \\ context\end{tabular} \\ \hline
\texttt{Dense}             & \begin{tabular}[c]{@{}c@{}}units=20,\\ activation='softmax'\end{tabular} & \begin{tabular}[l]{@{}l@{}}Classifies into 20 \\ meta-classes\\ using softmax \end{tabular} \\ \hline
\end{tabular}
\end{table}

For both fine-tuning and learning from scratch, the CG-CNN model effectively learns to distinguish each document from all others. The features developed through this process are well-suited for document classification, a task that is relatively easier than distinguishing arbitrary text blobs from each other. Both random forest and logistic regression classifiers learn significantly better with CG-CNN features compared to using standard Word2Vec embeddings. This improvement is particularly pronounced when CG-CNN is used for fine-tuning (rather than learning an embedding from scratch with a small unlabelled dataset). For our comparisons, once trained, CG-CNN's embeddings are utilized similarly to those from Word2Vec. For each document, the embeddings are averaged, and this average is then used as input to classifiers such as logistic regression (LR) and random forest (RF). The performance metrics outlined below underscore the effectiveness of CG-CNN. Using the LR classifier on CG-CNN's fine-tuned embeddings achieved a mean accuracy of \(75.36\%\) with a standard deviation of \(4.26\%\), demonstrating a significant improvement over the \(60.89\%\) baseline of Word2Vec. Additionally, a CG-CNN configuration extracting 50 dimensions (instead of 300) yielded a mean LR accuracy of \(71.19\%\). This configuration was used as a robustness check, effectively serving to validate the model's performance and to ensure there were no anomalies or bugs affecting the results. Using the RF classifier on CG-CNN's fine-tuned embeddings achieved a mean accuracy of \(70.76\%\) with a standard deviation of \(4.06\%\), demonstrating a significant improvement over the \(66.91\%\) baseline of Word2Vec. Additionally, using the LR classifier on CG-CNN's from-scratch embeddings achieved a mean accuracy of \(74.08\%\) with a standard deviation of \(1.06\%\) and the RF classifier achieved a mean accuracy of \(71.16\%\) with a standard deviation of \(1.13\%\).

\begin{figure}[t]
    \centering
    \includegraphics[width=\linewidth]{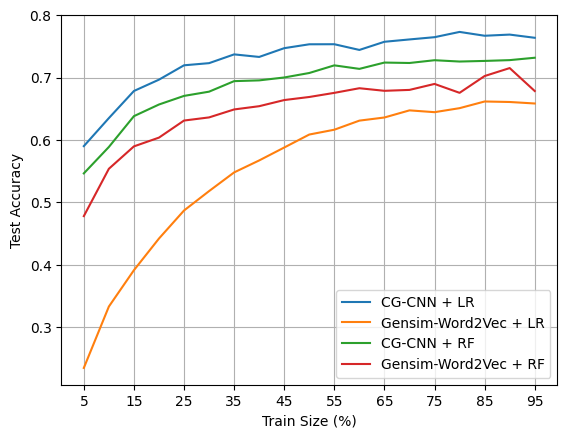}
    \caption{20 Newsgroups test accuracy vs. train size using Logistic Regression (a) and Random Forest (b) on the original and fine-tuned embeddings.
}
    \label{fig:text-combined}
\end{figure}

In a departure from the previously reported 50-25-25 split for the unsupervised training, supervised training, and testing sets, we now vary the proportions of the training and testing sets within the labeled data. Figure~\ref{fig:text-combined} illustrates the average of 10 runs for the classification accuracies of logistic regression (LR) and random forest (RF) classifiers using CG-CNN fine-tuning versus Word2Vec embeddings. This is shown as a function of the train-size proportion, demonstrating how variations in training data proportions affect model performance.

\section{Conclusions}
The Contextually Guided Convolutional Neural Network (CG-CNN) presents a robust alternative to traditional deep learning models by emphasizing the development of pluripotent features through contextual guidance and transfer learning. This paper builds on our previous work by showcasing CG-CNN’s broad applicability across various domains. Our experimental results, including applications to word embeddings for text classification with the 20 Newsgroups dataset and tactile sensing with the VibTac-12 dataset, demonstrate the adaptability of CG-CNN to both structured and unstructured data. Through its EM iterations, CG-CNN features progressively develop greater \emph{transfer utility}—a concept formulated and defined in this paper as the degree of usefulness when applied to new tasks. Detailed demonstrations on the classical XOR problem further provide deep insights into the model’s capacity to develop features with high transfer utility. CG-CNN features have shown superior classification accuracy compared to those from well-known deep networks like AlexNet, ResNet, and GoogLeNet, particularly in image-related tasks. Moreover, when fine-tuned through self-supervision on an unlabeled dataset, these features significantly outperform Word2Vec in text classification tasks on a downstream labeled test set. Overall, CG-CNN features provide substantial advantages in environments with limited data and requiring minimal model complexity.

\textcolor{black}{The current single-layer CG-CNN, while effective, is limited in the complexity of the features it can develop. Future enhancements might involve expanding to a multi-layer architecture, where each layer could be trained using local contextual guidance from the outputs of preceding layers. Harnessing the contextual information at higher levels within data remains a critical challenge for guiding the development of higher-level CNN layers and determining the contextual groups for their EM-based training. Additionally, integrating feedback guidance from higher to lower layers could significantly enhance the CG-CNN model. Addressing the issue of feature forgetting —particularly for older tasks— remains as an unsolved issue; mechanisms to prevent this forgetting are necessary to maintain model efficacy over time.} 

\section*{Acknowledgment}
This work was supported, in part, by the National Science Foundation under grants No. 2003740 and 2302537. The development of the semi-supervised learning algorithm benefited from the conceptual work involved in preparing NSF Grant No. 2435093, which is set to begin on November 1, 2024.

\bibliography{references}
\bibliographystyle{apalikeV2.bst}

\EOD
\end{document}